\documentclass[twocolumn, journal]{IEEEtran}

\usepackage{amssymb, graphicx, booktabs, longtable}
\usepackage{cite}
\usepackage{amsmath}
\usepackage{color}
\usepackage{multirow}
\usepackage{makecell}

\usepackage[final]{changes}
\definechangesauthor[name={lianghao han},color= blue]{LHH}
\usepackage{csquotes}
\usepackage{todonotes}
\usepackage[switch]{lineno}

\ifCLASSINFOpdf
\else
\fi

\begin{document}
%
\title{A Biologically Interpretable Two-stage Deep Neural Network (BIT-DNN) For Vegetation Recognition From Hyperspectral Imagery}
%
%
%

\author{Yue~Shi, Liangxiu~Han, Wenjiang~Huang, Sheng~Chang,  Yingying~Dong, Darren~Dancey,  Lianghao~Han

\thanks{ Yue Shi, Liangxiu Han, and Darren Dancey are with Department of Computing and Mathematics, Faculty of Science and Engineering, Manchester Metropolitan University, Manchester M1 5GD, UK.}
\thanks{ Wenjiang Huang, Sheng Chang and Yingying Dong are with Key Laboratory of Digital Earth Science, Aerospace Information Research Institute, Chinese Academy of Sciences, Beijing 100094, China.}
\thanks{Lianghao Han is with Department of Computer Science, Brunel University, UB8 3PH, UK}
\thanks {Corresponding author: L. Han (e-mail: l.han@mmu.ac.uk)}

}

\maketitle

\begin{abstract}
Spectral-spatial based deep learning models have recently proven to be effective in hyperspectral image (HSI) classification for various earth monitoring applications such as land cover classification and agricultural monitoring. However, due to the nature of "black-box" model representation, how to explain and interpret the learning process and the model decision, especially for vegetation classification, remains an open challenge. This study proposes a novel interpretable deep learning model -- a biologically interpretable two-stage deep neural network (BIT-DNN), by incorporating the prior-knowledge (i.e. biophysical and biochemical attributes and their hierarchical structures of target entities) based spectral-spatial feature transformation into the proposed framework, capable of achieving both high accuracy and interpretability on HSI based classification tasks. The proposed model introduces a two-stage feature learning process: in the first stage, an enhanced interpretable feature block extracts the low-level spectral features associated with the biophysical and biochemical attributes of target entities; and in the second stage, an interpretable capsule block extracts and encapsulates the high-level joint spectral-spatial features representing the hierarchical structure of biophysical and biochemical attributes of  these target entities, which provides the model an improved performance on classification and intrinsic interpretability with reduced computational  complexity. We have tested and evaluated the model using four real HSI datasets for four separate tasks (i.e. plant species classification, land cover classification, urban scene recognition, and crop disease recognition tasks). The proposed model has been compared with five state-of-the-art deep learning models. The results demonstrate that the proposed model has competitive advantages in terms of both classification accuracy and model interpretability, especially for  vegetation classification. 
\end{abstract}


%
\IEEEpeerreviewmaketitle

%
%
%
%

\section{Introduction}
\label{sec:1}
\IEEEPARstart{D}{eep} learning models have recently been used for hyperspectral image (HSI)-based vegetation monitoring applications, such as the agricultural monitoring \cite{RN4, RN5}, and ecological management \cite{RN8, RN9}. However, most of existing deep learning-based approaches have difficulty in explaining plant biophysical and biochemical characteristics due to the black-box representation of the features extracted from intermediate layers and the complex design of network architectures in a deep learning model \cite{RN11, RN10}. Therefore, the interpretability of deep models for HSI-based vegetation information recognition has become one of the most active research topics in the remote sensing community, which can enhance and improve the robustness and accuracy of models in the vegetation monitoring applications from the biological perspective of target entities \cite{RN13, RN14}.   \par

Some efforts on interpretable deep learning-based models in the remote sensing field have been made \cite{RN12}.  The visualisation of feature representations is the most direct way to improve the interpretability of a model \cite{RN34}. This type of methods adds an additional layer to visualize intermediate features or patterns, either through maximizing the score of a given unit in a pre-trained deep learning model or through inverting feature maps of an intermediate layer back to the input image \cite{RN30, RN53, RN28}. For example, Cai \textit{et al.} \cite{RN29} studied the spatial distribution and significant of the output of each layer, and proposed a multilayer visualisation approach to simultaneously visualize the sample distribution, the details of the subpixel level, the target units and labels hidden in the deep levels of airborne AVIRIS data and spaceborne Hyperion data. Another way to improve the interpretability of deep learning models is to construct a network architecture which can bring the network an explicit semantic meaning \cite{RN35, RN34}. For example, Lin \textit{et al.} \cite{RN36} proposed an unsupervised model, named as multiple-layer feature-matching generative adversarial networks (MARTA GANs), to explore and extract the representation of unlabelled data during the learning processes. In this model, a generative model was used to integrate local and global features, and a discriminative model was set to learn better spectral representations from HSI images. \par

Although existing researches are encouraging, plant-specific biophysical and biochemical attributes and their hierarchical structures, which provide the most direct evidence in indicating plant type and growth state, are still hard to be explained in the learning process. Here, the biophysical and biochemical attributes in HSI data refer respectively to foliar components' information (e.g.  nutrient and pigment level) and plant's structural information (e.g. the leaf area and angle) hidden in the spectral domain and the hierarchical structure. They represent the biological phenotype of the biochemical and biophysical components integrated in specific plant species in the joint spectral-spatial domains. The model interpretability on the hierarchical structure of these attributes is a key factor to measure the intelligence degree of a deep learning model understanding plant species and growth states in a biological way. However, the complexity and diversity in the reflectance radiation of plant canopies make the biological interpretability of a deep learning model challenging. On one hand, unavoidable spectral-spatial perturbations and redundancies in HSI data always cause difficulty in accurately representing  the features  of intermediate layers \cite{RN23}. On the other hand, it is hard to capture a hierarchical biological relationship between the high-level features produced by deeper layers \cite{RN26}. \par

Generally speaking, a well-designed interpretable model for HSI-based vegetation information recognition needs to deal with two issues: 1) how to extract the interpretable features that are associated with biological attributes of target entities, 2) how to represent the hierarchical structure of biophysical and biochemical attributes of target entities. To address these issues, in this study, a novel biological interpretable two-stage deep neural network (BIT-DNN) model is designed to achieve  accurate recognition on plant information from HSI data. Emphasizing on the biophysical and biochemical representation, this model is composed of two stages. The first stage is designed for low-level spectral feature extraction with enhanced biochemical and biophysical representations, based on the prior-knowledge multi-band spectral transformation (i.e. vegetation indices approach) for boosting the representation of vegetation biochemical or biophysical properties in the applications of vegetation properties retrieval and vegetation classification. The second stage is designed for characterizing the relationship and the hierarchical structure of high-level joint spectral-spatial features by integrating the spatial texture information with the spectral features extracted from the first stage. The contribution of this study lies in two-fold: 1) a novel two-stage deep learning model for accurate vegetation classification from HSI data; 2) Integration of the prior-knowledge based spectral-spatial analysis into the deep learning process, which can improve the performance of the proposed deep learning model for extracting features associated with the biological attributes of vegetation entities captured by HSI data, and enable improved interpretability of decision making. \par

The rest of this paper is organised as follows: Section \ref{sec:2} provides an overview of related work on interpretable deep learning models for HSI image classification; Section \ref{sec:3} presents our proposed interpretable deep neural network (BITS-DNN) for HSI-based vegetation information recognition; Section \ref{sec:4} introduces the criteria of interpretability assessment; Section \ref{sec:5} describes the experimental evaluation; Section \ref{sec:discussion} discuses the results and interpretability; \ref{sec:6} concludes the work.

\section{Related work on interpretable deep learning models for HSI image classification}
\label{sec:2}

The interpretability of a deep learning model should be considered in the life cycle of data science: date collection, pre-processing, data modelling, and post hoc analysis \cite{RN71}. The intrinsic interpretability of deep learning models on HSI data should address the representation of the reflection and radiation characteristics of ground entities. Existing deep learning models enhance their interpretability mainly from three aspects including: 1) Pre-model interpretability, 2) In-model interpretability, and 3) Post-model interpretability (post hoc analysis). \par
More specifically, the pre-model interpretability enhancement, which is prior to the main model construction stage, mainly focuses on enhancing the biological attributes of ground entities hidden in the HSI data. Two of the most popular approaches are data description standardization and explainable feature enhancement. For example, Gao \textit{et al.} \cite{RN79} proposed a dimensionality reduction method to explore the spectral and spatial characteristics of HSI data, this approach improved the representation of hyperspectral patch alignments in the main model, and performed well on the small sample learning. Ribeiro \textit{et al.} \cite{RN73} transformed original three colour channels into an interpretable dataset with a tensor representing the potential shapes or texture attributes of target objectives.\par

The in-model interpretability enhancement refers to using the causality or physical constraints on the main model to extract interpretable features with explicit semantic meaning. For instance, Paoletti \textit{et al.}\cite{RN37} developed a deep capsule network (a convolutional neural network (CNN) based model) for HSI classification tasks in order to better model the hierarchical relationships of features. Through the exploitation of the correlation of spectral-spatial features, this approach added the so-called "capsules" structures to a CNN network. The capsules allowed efficient handling of the high level complexity of entities, including their spatial positions in the image, associated spectral signatures and potential transformations. Although such approaches can improve the interpretability, they could always make these neural networks deeper. Thus, a large number of filters are added into the network architecture, leading to the vanishing gradient problem and the limited performance of activations and gradients in the training progress \cite{RN39, RN38}. From this aspect, the prior knowledge-based feature enhancement or encoding technology is an efficient way to uncover the discriminate spectral-spatial characteristics hidden in the raw hyperspectral images \cite{RN40}. For instance, in order to formalize and exploit the knowledge of automatic urban objects identification, Forestier \textit{et al.} \cite{RN41} proposed a knowledge-based deep learning model for urban object detection. In comparison with traditional CNN-based approaches, this model provided a better performance on the interpretation of HSR images through mapping the territory automatically. Li \textit{et al.} \cite{RN42} proposed an LiDAR (Light Detection and Ranging) technology based deep learning model to classify forested landslides, in which the prior-knowledge of object features was manually integrated into feature extraction layers. As a result, the output features from the intermediate layers provided interpretable information for the geological characteristics of the target landslide, which subsequently achieved a better performance for the forested landslide classification in steep and rugged terrain.\par

The post-model interpretability enhancement, which is generally decoupled from the main model, refers to explaining the representations of the intermediate outputs. Since Zhao \textit{et al.} \cite{RN32} introduced a principle component analysis (PCA) based activation function for optimizing deep spectral-spatial features collected from the HSI classification framework, various CNN-based interpretable deep learning models have been developed, focusing on the exploration and interpretation of the spectral-spatial pattern of target entities using post-interpretable approaches. For instance, Yang \textit{et al.} \cite{RN31} concatenated the pixelwise spectral-spatial features and visualized the features of fully connected layers in order to extract and explain the contributions and representations of intermediate outputs for the final classification. Mou \textit{et al.} \cite{RN53} explored the interpretation of the training process in an unsupervised fashion, they proposed an encoder-decoder paradigm, in which the significant information of the input HSI patches was extracted and represented in a lower dimensional space via a CNN encoder. Among these methods, the visualisation-based approach and the interpretable activation optimization are two most popular post-model interpretability approaches, while the visualisation-based approach is the most direct way to explore the high-level representations of the spectral information hidden in the deeper layers. Hu \textit{et al.} \cite{RN30} provided a detailed comparison of these two different ways for extracting and visualizing image features from different layers and encoding dense features at multiple scales into global features.\par         
However, most of the existing interpretable deep learning approaches were designed based on the statistical properties of the sample space \cite{RN45, RN44}. Thus, the learning process was modelled as a set of joint probability density functions, and a large number of high-quality labelled training data was required. These methods neglect the biophysical and biochemical attributes hidden in the redundancy information of the HSI data, which makes the classification performance highly depend on the scale and quality of labelled samples. Moreover, the effect of mixed-pixels, which may degrade the intra-class variability and exaggerate the inter-class similarity and produce feature interferences during the learning process, were often not fully considered \cite{RN47, RN46}. Therefore, most existing deep learning approaches often have a poor interpretability for high-level features of HSI data and the salt and pepper noises on the final classifications \cite{RN49}. \par

\section{The proposed method: a biologically interpretable two-stage deep neural network (BIT-DNN)}
\label{sec:3}
\begin{figure}[!t]   %
\centering  %
\includegraphics[width=3.5in]{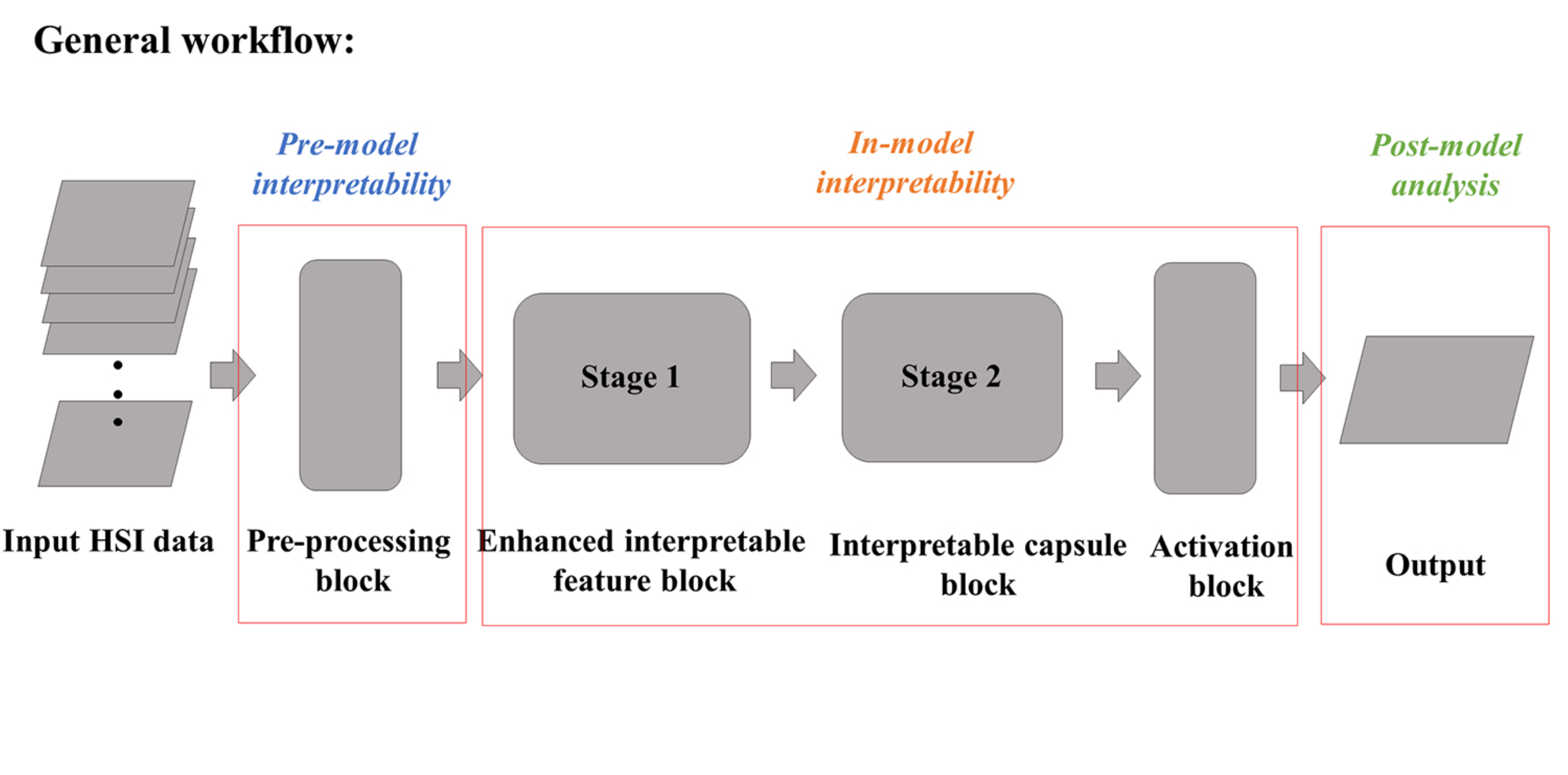}   %
\caption{The high-level system overview of the proposed biologically interpretable two-stage spectral-spatial deep neural network.}    %
\label{fig:1}  %
\end{figure}

In this study, we consider HSI data as a data cube, $\mathbf{X} \in \mathbb{R}^{H \times W \times B}$ with a size of $H \times W \times B$, where \textit{H}, \textit{W}, and \textit{B} are the height, width, and spectral bands of the original data cube, respectively, and each pixel comprises an individual spectral signal with \textit{B} spectral bands. We propose a novel deep learning framework, a biologically interpretable two-stage deep neural network (BITS-DNN) to deal with the HSI-based vegetation information recognition. In the model, we introduce a serial two-stage feature learning architecture, as shown in Fig.\ref{fig:1}. This design can provide great benefits on exploring the feature transformation during the learning process, which will be explained below. The proposed BITS-DNN model consists of input, four core blocks (pre-processing block, enhanced interpretable feature block named as stage 1, interpretable capsule block named as stage 2 and activation block) and output. The detailed design of each core block is described below.  \par

\subsection{The pre-processing block}
Considering that the explicit biochemical and biophysical properties of the vegetation classes are in different band ranges \cite{RN50, RN51}, we split HSI images into 7 slices: blue ($< 515nm$), green ($515 nm - 600 nm$), red ($600 nm - 680 nm$), red-edge1 ($680 nm - 710 nm$), red-edge2 ($710 nm - 750 nm$), red-edge3 ($750 nm - 790 nm$), near infrared ($>790 nm$). 

\subsection{Stage 1 -- An enhanced interpretable feature block}  

\begin{figure}[!t]   %
\centering  %
\includegraphics[width=3.5in]{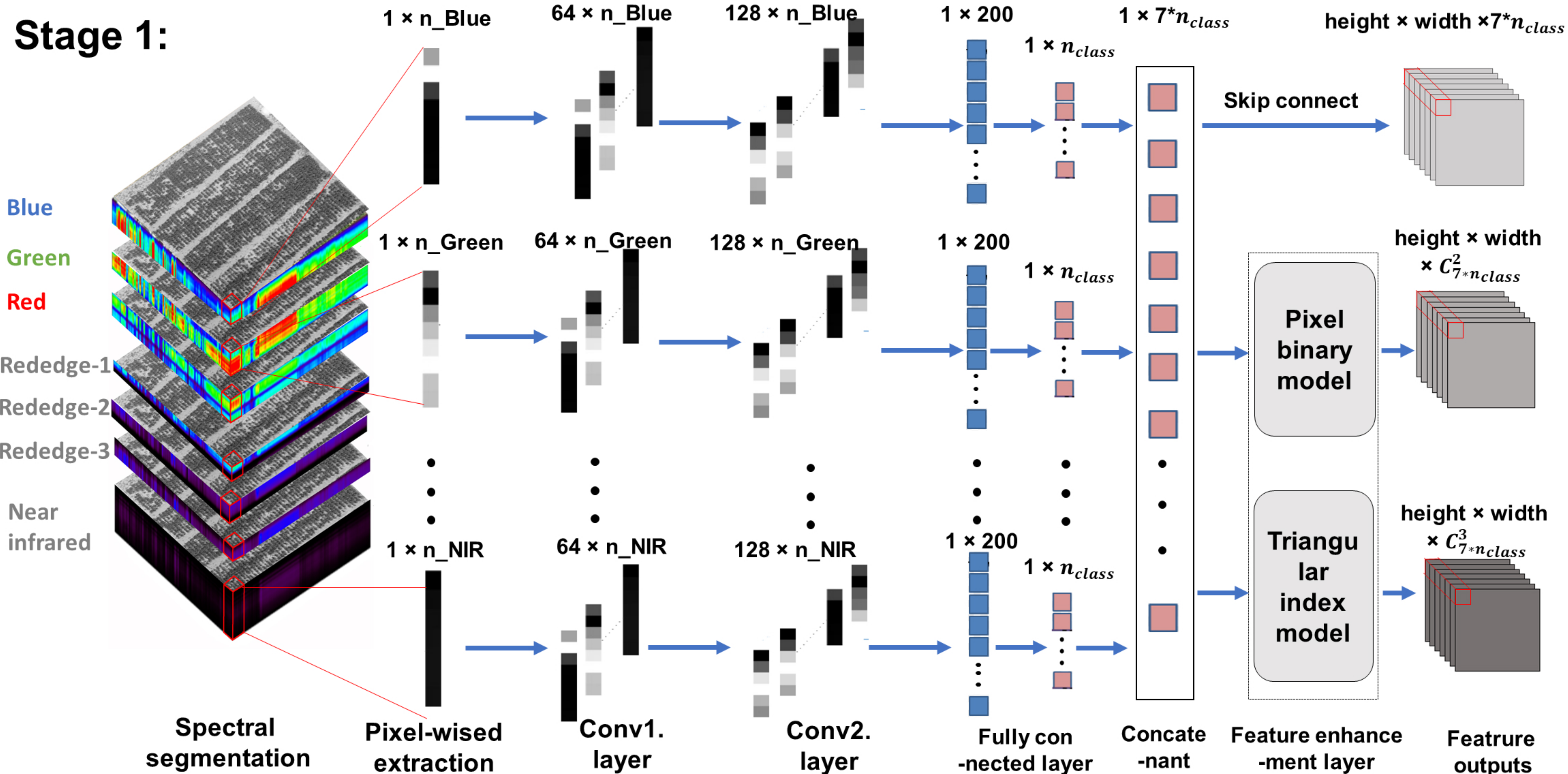}   %
\caption{The architecture of the enhanced interpretable feature block in the two-stage learning process, Stage 1. This stage is to extract and generate interpretable low-level spectral features.}    
\label{fig:2}  %
\end{figure}

The enhanced interpretable feature block named as Stage 1, is the first stage of feature learning and developed to extract and generate interpretable low-level features. Fig.\ref{fig:2} illustrates the architecture of this block. It includes two one-dimensional (1D) convolution layers, two fully connected (FC) layers (FC1 and FC2) and a feature enhancement layer. The two 1D convolution layers following with two fully connected layers are used for extracting the pixel-wised low-level features from the HSI slices, and the feature enhanced layer is designed to enhance the feature representations on vegetation biochemical and biophysical attributes and improve the model's interpretabilities.\par

As shown in Fig.\ref{fig:2}, the low-level features extracted from the fully connected layers after concatenation are further processed in the feature enhancement layer through mapping the low-level features into the multi-variate features using the pre-defined binary model and triangular index model, respectively. The proposed multi-variate features, which is inspired by the effectiveness of vegetation indices approaches\cite{RN59,RN60,RN62} in vegetation properties retrieval and vegetation classification, is designed to facilitate inter-comparisons of the low-level features and enhance the homogeneous terrestrial biochemical activities and canopy biophysical variations between different features. 
Introducing multi-variate feature transformations into deep learning models will benefit the enhancement of the feature representations on vegetation biochemical and biophysical properties, thus improving the vegetation classification performance and the interpretability of models.  \par
The detailed information about the convolutional layer, the fully connected layer and the feature enhancement layer in this block is described as follows:\par

\subsubsection{1D-convolutional layer}
In order to extract the radiation magnitude of the central bands and spectral texture between the central bands and their neighbour bands, two 1D-convolutional layers (i.e. \textit{conv1} and \textit{conv2} layer shown in Fig.\ref{fig:2}) are introduced into the proposed network. The pixel-wise features will be extracted through convolution operations on HSI images using a series of filters with various reception fields, formulated as follows:\par

\begin{equation}
C^j_p=\sum_{l=1}^{rm} W_{l}^j I_{l}^{p}
\end{equation}

where $rm$ is the size of the receptive field, $C_p^j$ is the layer output of the $p^{th}$ neuron with the $j^{th}$ filter, $W_{l}^j$ is the weight of the $l^{th}$ unit of the $j^{th}$ filter, and $I_{l}^p$ is the value of the $l^{th}$ pixel within the patch corresponding to the $p^{th}$ neuron. \par

\subsubsection{The fully connected layer}
Here, two fully connected layers (FC1 and FC2) are used to non-linearly integrate the 1D-convolutional features into the class-specific low-level features. For an $n$-class classification task, the output size of the second fully connected layer (FC2) for each band slice will be $n$.  After concatenation, the extracted low-level features for each pixel of an HSI image, denoted as $X_{out}^1$, has a size of $1 \times (m\cdot{n_{class}})$, where $m$ is the total number of band slices ($m=7$ in this study). For an HSI image with a spatial size of $H\times{W}$, the output of the second convolutional layer is $\mathbf{X}_{out}^1 \in \mathbb{R}^{H \times W \times{(7\cdot{n_{class}}})}$.

\subsubsection{Feature enhancement layer}
In order to facilitate inter-comparisons of the single low-level features and enhance the homogeneous terrestrial biochemical activities and canopy biophysical variations, the low-level features extracted from the HSI data, $X_{out}^1$, are further processed with two feature transformation models, the pixel binary index model and the triangular index model, borrowed from the concept of vegetation indices \cite{RN10}.\par

The binary index model is designed to conduct normalized differences between any two of the $7\cdot{n_{class}}$ low-level features, the transformed two-variate feature can be activated in the learning process if its componential low-level features in $X_{out}^1$ have homogeneous responses to a specific class. The binary index model is calculated as follows:
\begin{equation}
X_{out}^{2,k}=\frac{X_{out}^{1,i}-X_{out}^{1,j}}{X_{out}^{1,i}+X_{out}^{1,j}} \quad i\neq j \in [1,C^2_{7\cdot{n_{class}}}], k\in[1,C^2_{7\cdot{n_{class}}}]
\end{equation}
where $X_{out}^{2,k}$ is the output of the $k^{th}$ feature combination based on $i$ and $j$ features in in $X_{out}^1$. For each pixel of an HSI image, the output from the binary index model,  $X_{out}^{2}$, has a dimension of $1\times{C^2_{7 \cdot {n_{class}}}}$. For an HSI image with a spatial size of $H\times{W}$, the output from the binary combination model is $\mathbf{X}_{out}^2 \in \mathbb{R}^{H \times W \times {C^2_{7\cdot{n_{class}}}}}$.\par

The triangular index model is proposed to measure the geometric area of the triangular composed by any three of the $7\cdot{n_{class}}$ low-level features in the feature space. The transformed three-variate feature would be activated in our model if its componential three low-level features have consistent direction in the feature space. The triangular index model is defined as follows:
\begin{small}
\begin{equation}
\begin{split}
&X_{out}^{3,k}=\frac{|j-h|\times{(X_{out}^{1,i}-X_{out}^{1,h})}-|i-h|\times{(X_{out}^{1,j}-X_{out}^{1,h})}}{2} \\ &\quad i\neq j \neq h \in [1,C^3_{7\cdot{n_{class}}}], k\in[1,C^3_{7\cdot{n_{class}}}]
\end{split}
\end{equation}
\end{small}

where, $X_{out}^{3,k}$, is the output of the $k^{th}$ three-variate feature combination based on $i$,$j$ and $h$ features in $X_{out}^1$. The final output from the triangular index model for each pixel of an HSI image, denoted as $X_{out}^3$, has a dimension of $1 \times C^3_{7\cdot{n_{class}}}$. For an HSI image with a spatial size of $H\times{W}$, the output from the triangular index model is $\mathbf{X}_{out}^3 \in \mathbb{R}^{H \times W \times C^3_{7\cdot{n_{class}}}}$.\par

Finally, for each pixel of an HSI image, the output of the enhanced interpretable feature block, denoted as $X_{out} $, is the concatenation of $X_{out}^1, X_{out}^2, X_{out}^3$. For an HSI image with a spatial size of $H\times{W}$, the output of the feature maps from the first stage feature learning is $\mathbf{X}^{(1)}_{out}=\boldsymbol{[} \mathbf{X}^{1}_{out},\mathbf{X}^{2}_{out},\mathbf{X}^{3}_{out}, \boldsymbol{]} \in \mathbb{R}^{H \times W \times F_N}$, where $F_N$ is the total number of features for each pixel after feature concatenation, equals to $7\times{n_{class}}+C^2_{7\cdot{n_{class}}}+C^3_{7\cdot{n_{class}}}$ .

\subsection{Stage 2 -- An interpretable capsule block }
The interpretable capsule block is the second stage of feature learning and designed to better model the hierarchical structure of the biophysical and biochemical attributes of vegetation entities in order to achieve highly accurate classification and high interpretability. It consists of a 2D-convolutional layer, a capsule layer and a classification capsule layer (see Fig.\ref{fig:3}). Specifically, the output from the spectral-derived enhanced interpretable feature block (Stage 1) would be firstly input into a 2D-convolutional layer, in which the spatial texture information provided by the enhanced feature maps are integrated with their spectral-derived information, and then the jointly spectral-spatial features are outputted. Subsequently, the spectral-spatial feature maps would be fed into a capsule layer, where the spectral-spatial features would be encapsulated into a series of featured tensor as high-level features. Finally, in order to use these high-level features for classification, a class capsule layer is designed to output membership scores of certain feature vectors belonging to specific labels (classes). The detailed information about the layers in the block is described in the following subsections:
\begin{figure}[!t]   
    \centering  
    \includegraphics[width=3.5in]{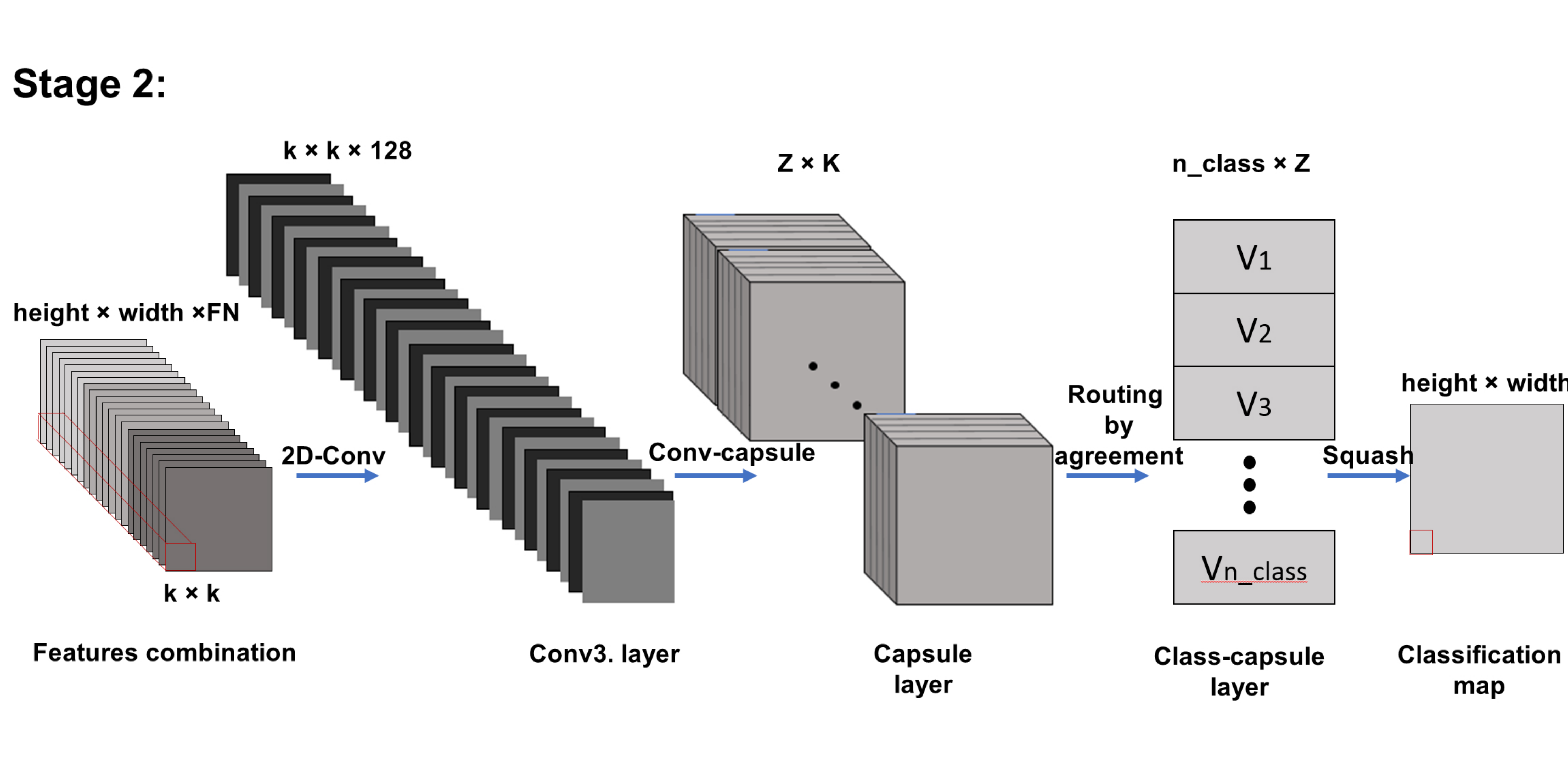}   
    \caption{The architecture of the interpretable capsule block in the two-stage learning processing, Stage 2. This stage is designed to better model the hierarchical structure of the biophysical and biochemical attributes of the target ground entities in order to achieve high accuracy and interpretability. }  
    
    \label{fig:3}  
\end{figure}

\subsubsection{2D-convolutional layer}
Here, we introduce a $2D$-convolutional layer (named as Conv3) to extract the spatial texture information. Since the convolution operation is applied on the spectral feature maps extracted from the enhanced interpretable feature block, it will generate joint spectral-spatial features. In other words, the convolutional operation can be regarded as a feature updating process to improve the representation of vegetation information through integrating spatial structural information and the pixel-wised spectral features. In this layer, we choose a total of $K^{(1)}$ filters, each of them has the same dimension of $k \times k \times F_N$. The input of this layer is the combined spectral feature sets from Stage 1,  $\mathbf{X}^{(1)}_{out}$, and its output is $\mathbf{O}_{out} \in \mathbb{R}^{H^{(1)}\times W^{(1)} \times K^{(1)}}$, where $H^{(1)}$ and $W^{(1)}$ are the height and weight of feature maps, respectively.\par

\subsubsection{Capsule layer}

To learn higher-level features from the lower-level spatial-spectral features, the convolution layers are often followed by a pooling layer. However, the pooling layer does not model the spatial hierarchical representation of features, and may lead to a poor performance in characterizing and detecting the pose relationship between target classes \cite{RN43}. Therefore, we do not use the pooling layer in this study. A previous study \cite{RN37} has shown that the capsule layer can be used for extracting and detecting the instantiation parameters (such as pose and orientation) of spectral-spatial features.  Therefore we add a capsule layers after the 2D convolutional layer to learn higher-level spatial-spectral features from the lower-level features extracted from Conv3 and preserve the hierarchical pose (translational and rotational) relationships between objects. The capsule layer has multiple capsules, and each capsule is a set of neurons that individually activate for various properties (such as position, size, deformation, texture etc.) of a type of objects. A capsule will output an activity vector rather than scalar-values, and use the length of the vector to represent the probability that an entity exists and its orientation to represent the properties of the entity.\par

In this study, the capsule layer comprises $Z$ convolutional capsules, and each capsule contains $K$ convolutional neurons with a kernel of  $\text{k} \times \text{k} \times K^{(1)}$. Its working mode is similar to CNN kernels. It takes the low-level spectral-spatial features, $\mathbf{O}_{out}$,  extracted from the Conv3 layer as input, and uses $\textbf{Z}$ capsules to detect the specific biophysical and biochemical features and learn the pose relationship between features. The $m^{th}$ capsule will apply its $K$ convolutional neurons over $\mathbf{O}_{out}$ to generate an output vector $\mathbf{u^{(m)}} \in \mathbb{R}^{K} = \boldsymbol{[} u^{(m)}_1,u^{(m)}_2,\cdots{},u^{(m)}_K \boldsymbol{]} $. The orientation of the output vector represents the spatial relationships of entities(or features), while its length represents whether the specific entities (or features) that a capsule is looking for exist. These encapsulated output vectors represent various attributes of the same entity, such as orientation, pose, size, biochemical or biophysical components in the HSI data. In addition, these feature tensors preserve much more information of biophysical and biochemical correlation relationships between the extracted spectral-spatial features and the target entities. To represent the length of the output vector as a probability value, it is often scaled down with a nonlinear squash function, formulated as follow:
\begin{equation}
\breve u_m=\frac{||u_m||^2}{1+||u_m||^2}\cdot\frac{u_m}{||u_m||}
\end{equation}
where $\breve u_m^{(l)}$ is the scaled activity vector. This function can be considered as a nonlinear activation function to make short vectors get shrunk to almost zero and long vectors get shrunk to a value slightly below 1.

The final output of the capsule layer, is denoted as $\mathbf{X}^{(2)}_{out} \in \mathbb{R}^{H^{(2)}\times W^{(2)}\times Z \times K}$, where $H^{(2)},W^{(2)}$ are the height and the width of feature maps, respectively.  \par

\subsubsection{Class-capsule layer}
The class-capsule layer is designed to connect the outputs of $Z$ capsules, $\mathbf{X}^{(2)}_{out}$, as the encoder units of target entities. In this work, the length of final encoder units is the number of classes, and the width of which is the number of the capsules (i.e. $Z$). For each input patch, the activity vectors will be encoded as the probability of belonging to corresponding entities. For this purpose, a dynamic routing algorithm proposed by Sabour \textit{et al.} \cite{RN43} is employed to connect the current layer with the previous capsule layer in order to iteratively update the parameters between these two layers. The aim of this step is to provide a well-designed learning process that not only connects the spectral information between capsules but also highlights the part-whole spatial correlation through reinforcing the connection coefficients between the different layers, and subsequently achieving accurate predictions. Mathematically, the encoder unit $\hat{u}_{n|m}^{(l)}$ in layer $l$ is formulated as:
\begin{equation}
\hat{u}_{n|m}^{(l)}=W_{m,n}^{(l)}\cdot\breve{u}_m^{(l-1)}+B_n^{(l)}
\end{equation}
where $\breve{u}_m^{(l-1)}$ is the $m^{th}$ capsule outputs in layer $l-1$, $B_n^{(l)}$ is the biases of the $n_{th}$ capsule in layer $l$ , and $W_{m,n}^{(l)}$ is a transformation matrix that connects the $m^{th}$ capsule output in layer $l-1$ with the $n^{th}$ capsule output in layer $l$. This formula allows the lower level capsules in layer $l-1$ to make prediction for superior capsules in layer $l$, improving the representation of the extracted features in biochemical-biophysical domain. Subsequently, a dynamic routing coefficient $c_{m,n}^{(l)}$ is introduced to reinforce the prediction agreement during the process of calculating the input $s_n^{(l)}$ of capsule n in layer $l$: 

\begin{equation}
s_n^{(l)}=\sum_m^{z^{(l-1)}}c_{m,n}^{(l)}\cdot\hat{u}_{n|m}^{(l)}
\end{equation}

where $c_{m,n}^{(l)}$ measures the contribution of the $m^{th}$ capsule in layer $l-1$ to activate the $n^{th}$ capsule in layer $l$, the sum of all the routing coefficient $c_{m,n}^{(l)}$ must be 1, and $c_{m,n}^{(l)}$ is obtained by:
\begin{equation}
c_{m,n}^{(l)}=\frac{e^{b_{m,n}}}{\sum_i^{z^{(l)}}e^{b_{m,i}}}
\end{equation}
where $b_{m,n}$ is the $log$ prior which indicates the correlation relationship between the $m^{th}$ capsule in layer $l-1$ and the $n^{th}$ capsule in layer $l$, it is initialized as 0 and is iteratively refined as follow:
\begin{equation}
b_{m,n}^l=b_{m,n}^{l-1}+v_n^{l-1}\cdot\hat{u}_{n|m}^{(l-1)}
\end{equation}
where $v_n^{l}$ is the activity vector of the capsule layer $l$, which can be calculated based on the function as follows: 
\begin{equation}
v_n^{l}=\frac{||s_n^{(l)}||^2}{1+||s_n^{(l)}||^2}\cdot\frac{s_n^{(l)}}{||s_n^{(l)}||}
\end{equation}
Conceptually, through the dynamic routing algorithm, the similar prediction from the capsule layer will be grouped, and subsequently capturing the robust prediction with clearer biochemical and biophysical meaning. Finally, the prediction performance can be calculated by the loss function ($L$) as follows:  
\begin{equation} \label{eq:Lm}
\begin{split}
L_{margin}=\sum_i^{n_{class}}T_i \max{(0,edge^+-||v_n^{l}||^2)}+\\ \mu(1-T_i)(\max(0,||v_n^{l}||-edge^-)^2)
\end{split}
\end{equation}
where $T_i$ is 1 when class $i$ is present in the data, otherwise is 0. The $edge^+$ and $edge^-$ works as the edge which forces the length of the $v_n^{l}$ into a set of small interval values to minimize the loss. Here the $edge^+$ is set to 0.9 and $edge^-$is set to 0.1, $\mu$ is a regularization parameter, which is set to 0.5 in order to avoid over-fitting and reduce the effect of the negative activity vectors. 

\subsection{The activation block}
The activation block consists of two fully connected layers which use the output activity vectors of the spectral-spatial feature from the capsule block as input to reconstruct the classification map, represented as $\tilde{\mathbf{Y}}\in\mathbb{R}^{H \times W}$. The proposed model mininizes the difference between the desired classification map from labelled data, $\bar{\mathbf{Y}}$, and the reconstructed classification map, represented as $\tilde{\mathbf{Y}}$. Meanwhile the model also encourages the capsule block to encode the most relevant instantiation parameters of the input data by mininizing the loss function defined in Eq.(\ref{eq:Lm}). The final loss function with Adam optimizer is defined as follow:
\begin{equation}
L_{end}=L_{margin}+\theta\cdot L_{reconstruction}
\end{equation}

where, $L_{reconstruction}=\lVert{\tilde{\mathbf{Y}}-\bar{\mathbf{Y}}\rVert}$ is the mean square error (MSE) loss between the desired outputs and the network's reconstructed (predicted) outputs, and $\theta$ is the learning rate which is set to 0.0005 to balance the weights between $L_{margin}$ and $L_{reconstruction}$ during the reconstruction of loss. \par

\section{Interpretability assessment methods}
\label{sec:4}
We assess the interpretability of the proposed model from three aspects: pre-model, in-model, and post-model interpretability.
\subsection{Pre-model interpretability}
In order to evaluate the pre-model interpretability of the proposed pre-processing block, two standard metrics, Shannon entropy and Dunn index, are used to measure and visualize the quality of labelled clusters. Shannon entropy measures the uncertainty and disorder within the information represented by the intermediate features; and the entropy for a class $C$ is defined as:

\begin{equation}
E(C)=-\sum_{i=1}^{M} p{(x_i)} \cdot log(p(x_i))
\end{equation}

where $p(x)$ is the contribution (or probability) of the feature $x_i$ to the class $C$. A low-entropy implies a high-concentration of the feature set within the same class. Dunn index is defined as the ratio of the minimum inter-class distance and the maximum intra-class distance, thus,

\begin{equation}
D_{index}=\min_{1\leq i\leq m}\min_{1\leq j\leq m,i \neq j}\frac{\sigma(C_i,C_j)}{\max\limits_{1\leq k\leq m} {\Delta k}}
\end{equation}

where $\sigma(C_i,C_j)$ is the inter-class distance defined by the L2-norm distance between the class center (mean feature sequence) of class $C_i$ and $C_j$, and $\Delta k = \max\limits_{x,y \in C_i}d(x,y)$ is the intra-class distance defined by the L2-norm distance between any two samples $x$ and $y$ with the same label. A larger Dunn index suggests a better clustering because it indicates a smaller intra-class distance or inter-class distance.\par
\subsection{In-model interpretability}
In our proposed model, the feature enhanced layers described in Section 3.2 and the capsule layers described in Section 3.3 are the intrinsically interpretable blocks, which consider the physical mechanism of the spectral combinations and the biological hierarchical interactions among the extracted spectral-spatial features, respectively. In this study, the results during the learning process will be stepwise outputted in order to evaluate the in-model interpretability.\par

\subsection{Post-model (Post Hoc) interpretability}
To evaluate the post model interpretability, auxiliary data are used to explain the biophysical or biochemical meanings of the intermediate features generated in the hidden layers of the model. This is decoupled from the main model. Thus, it is only used to evaluate the interpretability of the intermediate layers without affecting the performance of the main model. In this study, considering the difference of two typical applications, landcover classification and crop disease detection, we have selected two different types of auxiliary data for the post hoc analysis.\par
1)	Vegetation indices data \par
For land cover classification tasks, we have used vegetation indices data which are biochemical- and biophysical-associated. Vegetation indices designed to highlight a particular property of vegetation have proven to be sensitive to certain biological attributes (e.g. indices listed in Table \ref{table:1}). They are calculated by combinations of surface reflectance at two or more wavelengths. Here we chose 10 popular vegetation indices. In order to quantify the biological attributes of the intermediate features extracted in the deep layers, the coefficients of determination ($R^2$) between these features and the vegetation indices were calculated based on univariate correlation analyses.


\begin{table}[!t]
\renewcommand{\arraystretch}{1.3}
\caption{The biophysical- and biochemical-associated vegetation indices used in this study}
\label{table:1}
\centering
\resizebox{3.2in}{!}{
\begin{tabular}{cccc}
\toprule
Vegetation index & Relate to & Formula & Ref  \\ 
\midrule Normalized Difference \\Vegetation Index (NDVI) & Vegetation coverage & $\frac{R_760-R_560}{R_760+R_560}$ & \cite{RN70} \\
Photochemical Reflectance\\ Index (PRI) & Photosynthetic efficiency & $\frac{R_570-R_531}{R_270+R_531}$ & \cite{RN59} \\  
Red-edge Chlorophyll \\Index (CIred-edge) & Chlorophyll content & $\frac{R_760}{R_560}-1 $ & \cite{RN62}   \\
Normalized Difference \\Water Index (NDWI) & Water & $\frac{R_860-R_1240}{R_860+R_1240} $ & \cite{RN64} \\
Triangular Vegetation\\ Index (TVI) & Green LAI & $0.5[120(R_750-R_550)-200(R_670-R_550)]$ & \cite{RN60} \\
Structural Independent \\Pigment Index (SIPI) & Pigment content  & $\frac{R_800-R_445}{R_800+R_680}$ & \cite{RN67} \\
Plant Senescence \\Reflectance Index (PSRI) & Nutrient & $\frac{R_678-R_550}{R_750}$ & \cite{RN66} \\
Normalized Pigment Chlorophyll\\ ratio index (NPCI) & Chlorophyll density & $\frac{R_680-R_430}{R_680+R_430}$ & \cite{RN61} \\
Optimized Soil Adjusted \\Vegetation Index (OSAVI)  & soil background & $\frac{R_760-R_560}{R_760+R_560+0.16}$ & \cite{RN69} \\ 
\bottomrule
\end{tabular}
}
\end{table} 

2) Biological parameters data \par
For the crop disease detection, we employed measured biological parameters, including leaf area index (LAI), leaf chlorophyll content (CHL), leaf anthocyanin content (ANTH), nitrogen balance index (NBI), and percentile dry matters (PDM). They were synchronously measured at the same place where the HSI measurements were collected. In order to ensure the sample scale and spatial resolution of HIS data are consistent, a total of 72 sampling sites with $0.05 m \times 0.05 m$ subplots were set. The CHL, ANTH, and NBI were measured with Dualex Scientific sensor (FORCE-A, Inc. Orsay, France), a hand-held leaf-clip sensor designed to non-destructively evaluate the content of pigments and epidermal flavonol. For the LAI acquisition, LAI-2200 Plant canopy analyzer (Li-Cor Biosciences Inc., Lincoln, NE, USA) was used. For the PDM measurement, 10-12 leaves for each sampling subplot were weighed with an electronic balance (Haozhuang, Inc, Shanghai, China) and dried in an electric blowing drying oven (DGG-9240A, Senxin, Inc, Shanghai, China) over 10 hours. After drying, the percentile dry matter (PDM) of the leaves was calculated by the ratio of dry and fresh weight.\par
In order to find the linear correlation between model learned spectral features and these biological parameters, a correlation analysis is used. The coefficient of determination ($R^2$) is used to assess the interpretability of such features in the learning process.

\section{Experimental Evaluation}
\label{sec:5}
To evaluate the effectiveness of the proposed model, we have applied it to four real datasets (see Table \ref{table:2}) and have compared it with five state-of-the-art deep learning models for HSI classification and crop disease detection tasks. These models include 1) deep fully convolutional neural network (DFCNN) \cite{RN76}, 2) vectorized convolution neural networks (VCNNs) \cite{charmisha2018dimensionally}, 3) spectral-spatial convolutional neural network (SSCNN) \cite{mei2019spectral}, 4) spectral-spatial residual network (SSRN) \cite{RN77} and 5) the capsule network (CapsNet) \cite{RN37}. The detailed information of experimental configurations and evaluation is described below.  
\subsection{HSI data description}
Four HSI datasets including three public available datasets, Indian Pines (IP) dataset, Pavia centre (PC) dataset, University of Houston (UH) dataset and an experimentally measured Wheat Yellow Rust (WYR) dataset, are used for the evaluation and validation of the proposed model. Specifically, the IP dataset is used for testing our model on vegetation species classification, the PC dataset is used for the task of landcover types classification, the UH dataset is used for the task of urban scene recognition, and the WYR data is used for the task of crop diseases diagnosis. The detailed descriptions of these two datasets are presented as follows:\par
1)	IP dataset\par
The IP dataset was collected by the Airborne Visible and Infrared Imaging Spectrometer (AVIRIS) sensor in 1992, which covers different plant species in north-western Indiana, USA, and contains a total of 16 ground truth classes. This dataset involves 224 hyperspectral bands in the range of $400\:nm \sim 2500\:nm$ with a size of $145 \times 145$ pixels. The detailed information about the IP data set can be found in \cite{RN56}.\par
2) PC dataset\par
The PC dataset was collected by the ROSIS sensor, and covers a total of 9 landcover types in at Pavia, Northern Italy. This dataset involves 102 hyperspectral bands with a size of $1096 \times 715$ pixels. \par
3) UH dataset \par
The UH dataset was collected by the Compact Airborne Spectrographic Imager (CASI) over the urban areas with 15 labelled urban scene, Houston, USA. This dataset involves 144 hyperspectral bands in the range of $346 \:nm$ to $1046 \:nm$ with a size of $1905 \times 349$ pixels. \par
4)	WYR dataset \par
The WYR dataset was collected by the DJI S1000 UAV system (SZ DJI Technology Co Ltd., Gungdong, China) based on the UHD-185 Imaging spectrometer (Cubert GmbH, Ulm, Baden-Warttemberg, Germany) in 2018. This dataset involves the 125 bands from visible to near-infrared bands between $450 \: nm$ and $950 \:nm$ with a size of $16279 \times 14762$ pixels. All the images were obtained at a flight height of 30 $m$, with a spatial resolution close to $2 \: cm$ per pixel. Hyperspectral images were manually labelled based on the ground synchronization survey of the occurrence conditions of yellow rust.

\begin{table}[!t]
\caption{The Number of available samples in the IP, PC, UH, and WYR datasets} 
\label{table:2}
\centering
\resizebox{3.5in}{!}{
\begin{tabular}{ccccccccc}
\toprule
 & \multicolumn{2}{c}{IP dataset}   & \multicolumn{2}{c}{PC dataset}              & \multicolumn{2}{c}{UH dataset} & \multicolumn{2}{c}{WYR data set}                              \\
No. & Plant species          & Samples & Landcover types      & Samples              & Urban scence        & Samples  & \multicolumn{1}{c}{Crop status} & \multicolumn{1}{c}{Samples} \\ \hline
1   & Alfalfa                & 46      & Water                & 824                  & Healthy grass       & 921      & \multicolumn{1}{c}{Health}      & \multicolumn{1}{c}{10842}   \\
2   & Corn-notill            & 1428    & Trees                & 820                  & Stressed grass      & 746      & \multicolumn{1}{c}{Yellow rust} & \multicolumn{1}{c}{7682}    \\
3   & Corn-min               & 930     & Asphalt              & 816                  & Synthetic   grass   & 423      & \multicolumn{1}{c}{Others}      & \multicolumn{1}{c}{3613}    \\
4   & Corn                   & 237     & Self-Blocking Bricks & 808                  & Trees               & 514      &                                 &                             \\
5   & Grass/Pasture          & 483     & Bitumen              & 808                  & Soil                & 285      &                                 &                             \\
6   & Grass/Trees            & 730     & Tiles                & 1260                 & Water               & 468      &                                 &                             \\
7   & Grass/Pasture-mowed    & 28      & Shadows              & 476                  & Residential         & 682      &                                 &                             \\
8   & Hay-windrowed          & 478     & Meadows              & 824                  & Commercial          & 589      &                                 &                             \\
9   & Oats                   & 20      & Bare Soil            & 820                  & Road                & 574      &                                 &                             \\
10  & Soybeans-notill        & 972     & \multicolumn{1}{c}{} & \multicolumn{1}{c}{} & Highway             & 216      &                                 &                             \\
11  & Soybeans-min           & 2455    & \multicolumn{1}{c}{} & \multicolumn{1}{c}{} & Railway             & 546      &                                 &                             \\
12  & Soybeans-clean         & 593     & \multicolumn{1}{c}{} & \multicolumn{1}{c}{} & Parking Lot 1       & 318      &                                 &                             \\
13  & Wheat                  & 205     & \multicolumn{1}{c}{} & \multicolumn{1}{c}{} & Parking Lot 2       & 257      &                                 &                             \\
14  & Woods                  & 1265    & \multicolumn{1}{c}{} & \multicolumn{1}{c}{} & Tennis Court        & 108      &                                 &                             \\
15  & Bldg-Grass-Tree-Drives & 386     & \multicolumn{1}{c}{} & \multicolumn{1}{c}{} & Running Track       & 45       &                                 &                             \\
16  & Stone-steel   towers   & 93      & \multicolumn{1}{c}{} & \multicolumn{1}{c}{} &                     &          &                                 &                             \\
17  & Background             & 10776   & \multicolumn{1}{c}{} & \multicolumn{1}{c}{} &                     &          &                                 &                             \\\bottomrule
\end{tabular}
}
\end{table}

\subsection{Evaluation metrics}
To measure the effectiveness of the proposed model, we have used the following metrics: the overall accuracy (OA) and average accuracy (AA) \cite{RN81}, sensitivity and specificity \cite{RN80}, kappa coefficient \cite{RN82} and execution time. 
In addition, McNemar's chi-squared ($\chi^2$) test \cite{Dietterich1998} was used to evaluate the statistical significance of the accuracy differences between the proposed method and the state-of-art deep learning models. The McNemar's statistic test is described as

\begin{equation}
\chi^2=\frac{{(|f_{12}-f_{21}|-1)}^2}{f_{12}+f_{21}}
\end{equation}

where $f_{12}$ stands for the number of pixels classified correctly by the first classifier and wrongly by the second classifier and $f_{21}$ stands for the number of pixels classified correctly by the second classifier and wrongly by the first classifier.

\subsection{Model evaluation}
To evaluate the proposed model, we set up four experiments on four widely used datasets: (1) Plant species classification using the IP dataset, (2) Land cover classification using the PC dataset, (3) Urban scene recognition using the UH dataset and (4) Crop stress detection using the WYR dataset. We test our model on both large samples and small samples. For the large sample testing, two-thirds of all labelled pixels from the whole dataset were randomly selected as the training dataset, and the remaining pixels were used as the testing dataset. For the small sample testing, $10\%$ to $80\%$ of all labelled pixels were randomly chosen as the training set, respectively, and $20\%$ of the pixels in the dataset are chosen as the test set. The small sample testing was used to asses the effect of sample size on the model performance, while the size effect of input patches was investigated through choosing four different size configurations, $5 \times 5$ pixels, $7 \times 7$ pixels, $9 \times 9$ pixels and $11 \times 11$ pixels.

We compared the proposed model with five state-of-the-art deep learning models, DFCNN \cite{RN76}, VCNNs \cite{charmisha2018dimensionally}, SSCNN \cite{mei2019spectral}, SSRN \cite{RN77} and CapsNet \cite{RN37} in these experiments.

\subsubsection{Experiment One}Plant species classification using Indian Pines (IP) dataset.\par
In this experiment, we evaluate the performance of the proposed method on the plant species classification using the IP dataset. \par
Firstly, we test the proposed model on large samples with $2/3$ of all labelled pixels as its training set. Table \ref{table:IP-AA} lists the results of overall accuracy (OA) and standard deviation (STD) of the proposed approach and its five competitors with four different size configurations for input patches. It can be found that the proposed approach consistently outperforms its competitors in terms of average classification accuracy. Compared with the best model among the five competitors, the proposed approach achieves an improvement up to $5.92\%$ for $5 \times 5$, $3.84\%$ for $7 \times 7$, $4.86\%$ for $9 \times 9$ and $5.75\%$ for $11 \times 11$ patch size configurations, respectively. 
Meanwhile the standard deviations produced by the proposed approach are substantially lower than those produced by the competitors for most of cases. Therefore, the proposed model has a lower uncertainty in the learning process. It can also be found that all the models are not very sensitive to these four patch sizes in terms of OA. The optimal patch size is $7 \times 7$ for the proposed model, DFCNN, SSCNN, SSRN and CapsNet, $9 \times 9$ for VCNNs.  In terms of execution time, the proposed model does not show advantages over its competitors. As shown in Table \ref{table:IP-trainsample}, the proposed model is only faster than SSRN, this may be due to a large number of model parameters and the relevant complexity of the network architecture.  \par

\begin{table}[!t]
\caption{The overall accuracy (OA) and standard deviation (STD) of plant species classification on the IP dataset from our proposed model and its five competitors with four different size configurations for input patches.(Note: the best size configuration for each model is highlighted in bold.)}  

\label{table:IP-AA}
\centering
\resizebox{3.4in}{!}{
\begin{tabular}{ccccccccc}
\toprule
\multicolumn{1}{l}{} & \multicolumn{2}{c}{5 $\times$ 5}            & \multicolumn{2}{c}{7 $\times$ 7}                             & \multicolumn{2}{c}{9 $\times$ 9}                             & \multicolumn{2}{c}{11 $\times$ 11}                           \\
Class                & \multicolumn{1}{l}{OA(\%)} & Std(\%) & \multicolumn{1}{l}{OA(\%)} & Std(\%)                  & \multicolumn{1}{l}{OA(\%)} & Std(\%)                  & \multicolumn{1}{l}{OA(\%)} & Std(\%)                  \\ \hline
Proposed             & 96.57       & 3.41    & \textbf{98.05}      & 2.45        & 97.24           & 3.42            & 97.14             & 3.57 \\
DFCNN                & 39.31       & 7.92    & \textbf{46.32}      & 5.13        & 44.38           & 8.48            & 42.75             & 7.35  \\
VCNNs                & 69.61       & 5.56    & 70.28               & 5.45        & \textbf{72.03}  & 6.43            & 70.19             & 7.03                     \\
SSCNN                & 82.15       & 5.28    & \textbf{88.80}      & 4.89        & 85.86           & 4.80            & 81.97             & 4.60                      \\
SSRN                 & 90.13       & 5.40    & \textbf{92.69}      & 5.95        & 92.38           & 4.41            & 91.39             & 4.30                      \\
CapsNet              & 90.02       & 3.56    & \textbf{94.21}      & 4.52        & 91.00           & 2.05            & 90.27             & 3.76                     \\  \bottomrule
\end{tabular}
}
\end{table}

\begin{table}[!t]
\caption{Execution time and the number of model parameters of the proposed model and its competitors based on the optimal patch size configuration for IP data} 
\label{table:IP-trainsample}
\centering
\resizebox{3.2in}{!}{
\begin{tabular}{ccccccc}
\toprule
          & Proposed & DFCNN  & VCNNs & SSCNN  & SSRN  & CapsNet  \\\hline
Parameters & 91057    & 102052 & \textbf{41862} & 113548 & 74956 & 83761   \\
Time(s)    & 355.6    & 310.8  & \textbf{217.4} & 345.8  & 371.6 & 289.2 \\\bottomrule
\end{tabular}
}
\end{table}

\begin{table}[!t]
\caption{Plant species classification of the IP dataset from six models with the optimal patch size configuration for each model.}

\label{table:3}
\centering
\resizebox{3.2in}{!}{
\begin{tabular}{cccccccc}
\toprule
class                  & Proposed & DFCNN & VCNNs & SSCNN & SSRN  & CapsNet \\ \hline
Alfalfa                & \textbf{98.21}     & 62.54   & 80.17   & 75.24    & 98.11   & 95.41    \\
Corn-notill            & \textbf{98.43}     & 56.25   & 75.28   & 92.27   & 95.25    & 95.85    \\
Corn-min               &\textbf{98.75}     & 41.25   & 82.17   & 88.12    & 96.71    & 96.5      \\
Corn                   &\textbf{97.28}     & 82.55   & 69.24   & 84.55    & 93.65    & 96.11    \\
Grass/Pasture          & \textbf{98.81}    & 92.17  & 88.32   & 86.54    & 95.00       & 98.31    \\
Grass/Trees            &\textbf{96.65}     & 31.54   & 67.21   & 72.51    & 92.95    & 94.25   \\
Grass/Pasture-mowed    &\textbf{99.18}     & 88.45   & 92.81  & 89.25    & 94.41    & 98.46    \\
Hay-windrowed          & \textbf{99.54}     & 13.98   & 67.10    & 56.58    & 96.24   & 99.43   \\
Oats                   & \textbf{98.13}     & 55.91   & 68.25   & 97.21  & 96.25   & 97.47    \\
Soybeans-notill        & \textbf{99.44}     & 88.28   & 66.11   & 95.18  & 96.21   & 95.35    \\
Soybeans-min           & \textbf{98.78}     & 55.42   & 81.25   & 93.51   & 95.55   & 96.81    \\
Soybeans-clean         & \textbf{99.42}     & 91.24  & 72.15   & 96.66  & 93.88    & 93.65    \\
Wheat                  & \textbf{99.11}     & 97.41  & 91.24  & 92.45   & 94.51    & 97.48    \\
Woods                  & \textbf{97.25}     & 49.28   & 92.31  & 96.24   & 95.25    & 95.51    \\
Bldg-Grass-Tree-Drives & \textbf{97.68}     & 88.44   & 82.28   & 90.38    & 93.21    & 96.58    \\
Stone-steel towers     & \textbf{97.24}      & 82.38   & 93.35  & 92.12   & 95.35    & 95.46   \\
Background             & \textbf{97.67}      & 33.45   & 65.71   & 88.14    & 90.37    & 92.41    \\ \hline
OA(\%)                 & \textbf{98.05}	    & 46.32	  & 72.03	& 88.8	  & 92.69	  & 94.21   \\
AA(\%)                 & \textbf{98.33}     & 65.28   	& 78.48	 & 87.61  & 94.84  & 96.11   \\
Kappa                  & 0.853             & 0.498 & 0.781 & 0.825 & 0.834 & 0.811   \\ \bottomrule

\end{tabular}
}
\end{table}

Secondly, we analysed the classification performance of the proposed model on each plant type in details. Table \ref{table:3} provides a detailed comparison on the accuracy of each of 17 classes between the proposed model and five competitors with their own optimal patch size configuration.  Compared with its competitors, the proposed model achieves the best classification performance on all classes. In terms of the classification accuracy, the proposed model ranks in first place, reaching the OA of $98.05\%$ and AA of $98.33\%$, with a Kappa value of 0.85. The CapsNet has the second best classification performance (OA=$94.21\%$, AA=$96.11\%$, Kappa=$0.81$). Following the CapsNet, the SSRN achieves an OA of $92.69\%$ and an AA of $94.84\%$ with a Kappa value of $0.83$. In contrast, the DFCNN and VCNNs which do not consider the joint spectral-spatial information, perform much worse, especially on identifying the categories of Hay-windrowed and Grass/Trees.

To evaluate the statistic significance of the accuracy differences between two-paired models, we have also conducted McNemar's chi-squared ($\chi^2$) test, as shown in Table \ref{table:t1_stest}.  The results show that the improvement of our proposed model in overall accuracy against the best model of its five competitors, CapsNet, is statistically significant, with $\chi^2=14.32(p \leq 0.05)$ (see the last column and last row of Table \ref{table:t1_stest}). The differences in overall accuracy between the proposed model and all other competitors are statistically significant, with $\chi^2=39.56(p \leq 0.01)$ for the proposed vs. DFCNN, $\chi^2=24.46(p \leq 0.01)$ for the proposed vs. VCNNs, $\chi^2=23.19(p \leq 0.01)$, $\chi^2=14.21(p \leq 0.05)$ for the proposed vs. SSRN. This further confirms that the classification accuracy improvement of the proposed model on IP dataset over its competitors is statistically significant.

\begin{table}[!t]
\caption{McNemar's chi-squared ($\chi^2$) test with associated probability values ($P$) for evaluating the statistic significance of accuracy differences in class-wise predictions between paired models on the IP dataset. $***$ = $P \leq 0.01$, $**$ = $P \leq 0.05$, $*$ = $P \leq 0.1$.}
\label{table:t1_stest}
\centering
\resizebox{3.4in}{!}{
\begin{tabular}{cccccc}
\toprule
Class           & \makecell[c]{Proposed \\ vs. DFCNN}     & \makecell[c]{Proposed \\ vs. VCNNs}     & \makecell[c]{Proposed \\vs. SSCNN}     & \makecell[c]{Proposed \\vs. SSRN}     & \makecell[c]{Proposed \\vs. CapsNet} \\\hline
Alfalfa         & 30.75***              & 25.56**               & 24.72**               & 22.19**               & 17.40**               \\
Corn-notill     & 41.98***              & 37.58***              & 29.88***              & 25.10***              & 19.09**              \\
Corn-min        & 36.83***              & 27.66***              & 27.50***              & 20.61**               & 17.21**              \\
Corn            & 22.21**               & 21.17**               & 15.21**               & 12.50**               & 12.46**              \\
Grass/Pasture   & 33.23***              & 27.66***              & 20.91**               & 18.22**               & 17.10**                \\
Grass/Trees     & 33.58***              & 26.11***              & 23.59***              & 18.75**               & 14.93**                \\
Grass/Pasture-mowed    & 40.20***       & 30.53***              & 27.24***              & 20.40**               & 15.76**                \\
Hay-windrowed          & 43.17***       & 33.97***              & 26.64***              & 24.02***              & 17.13**                \\
Oats                   & 40.02***       & 34.34***              & 28.11***              & 21.76**               & 20.38***               \\
Soybeans-notill        & 36.99**        & 27.50***              & 27.45***              & 22.03**               & 20.18***               \\
Soybeans-min           & 26.02**        & 25.25**               & 22.35**               & 21.43**               & 15.81**                \\
Soybeans-clean         & 36.53***       & 28.37***              & 27.10***              & 20.86**               & 20.18***               \\
Wheat                  & 27.99**        & 23.45**               & 19.19***              & 19.38**               & 18.51**                \\
Woods                  & 26.72**        & 19.35**               & 16.81**               & 18.48**               & 17.86**                \\
Bldg-Grass-Tree-Drives & 28.85***       & 25.81***              & 19.41***              & 16.21**               & 11.83**                \\
Stone-steel towers     & 12.46**        & 10.32*                & 9.95*                 & 8.63*                 & 6.91*                   \\
Background             & 11.18*         & 11.76*                & 16.94*                & 15.97*                & 10.24*                 \\  \hline
Overall                & \textbf{39.56***}   & \textbf{24.46***}     & \textbf{23.19***}    & \textbf{14.21**}     & \textbf{14.32**}     \\ \bottomrule
\end{tabular}
}
\end{table}

\begin{figure}[!t]   %
    \centering  %
    \includegraphics[width=3.5in]{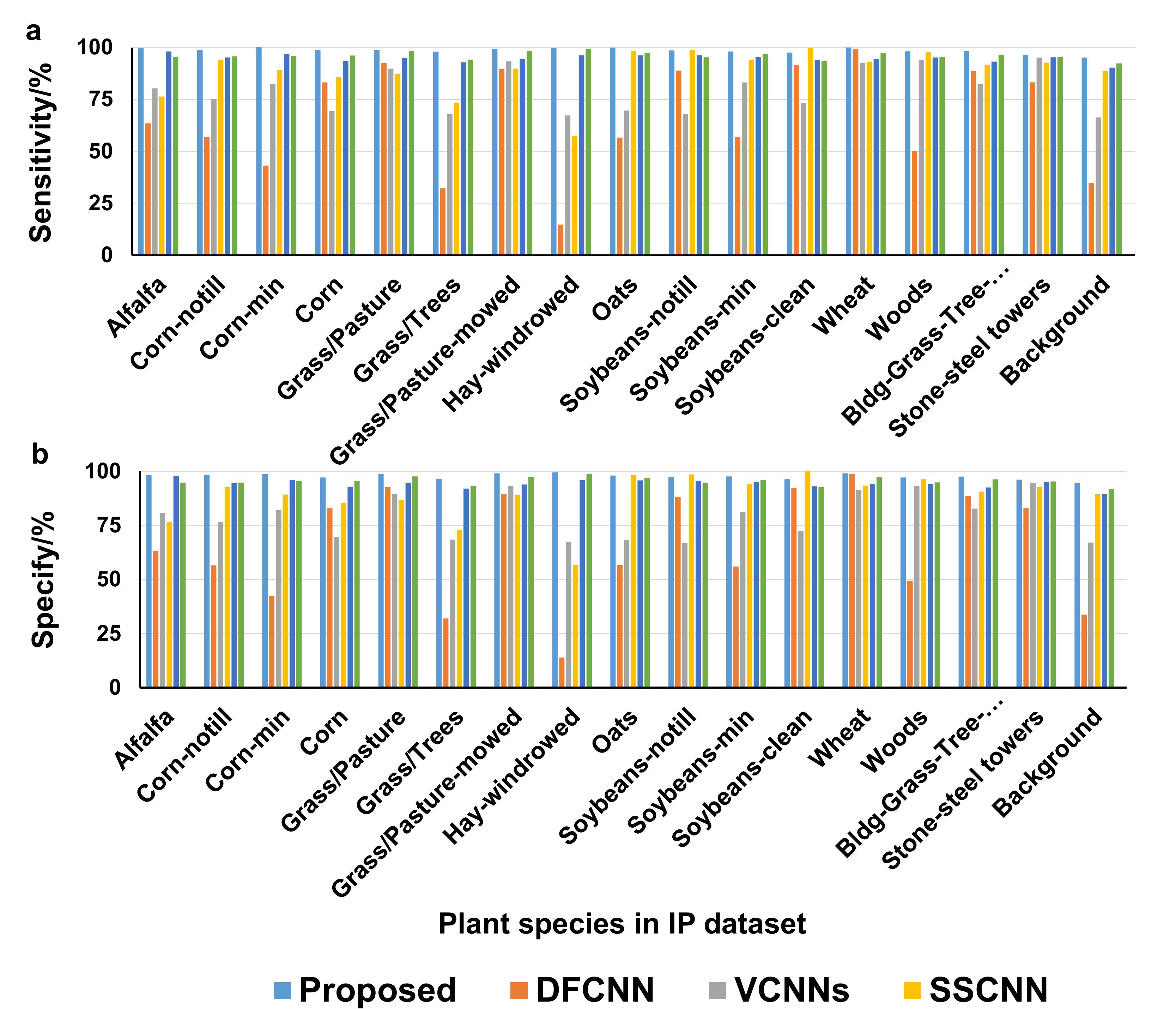}   
    \caption{A comparison of the sensitivity and specificity of each of 17 classes in the multi-class classification of the IP dataset from the models based on the optimal patch size configuration for each model (a) the sensitivity of each vegetation class, (b) the specificity of each vegetation class.}    %
    \label{fig:4}  %
\end{figure}

Fig. \ref{fig:4} illustrates a detailed comparison of the sensitivity and specificity of each class between six models on the IP dataset. It can be observed that the proposed approach achieves the highest sensitivity and specificity on identifying all of the vegetation categories except the soybeans-clean, which indicates that the proposed approach outperforms all the competitors in reducing the leakage and misclassification. Overall, the proposed approach outperforms the competitors with a considerable improvement in the plant species classifications. \par

\begin{figure}[!t]   %
    \centering  %
    \includegraphics[width=3.5in]{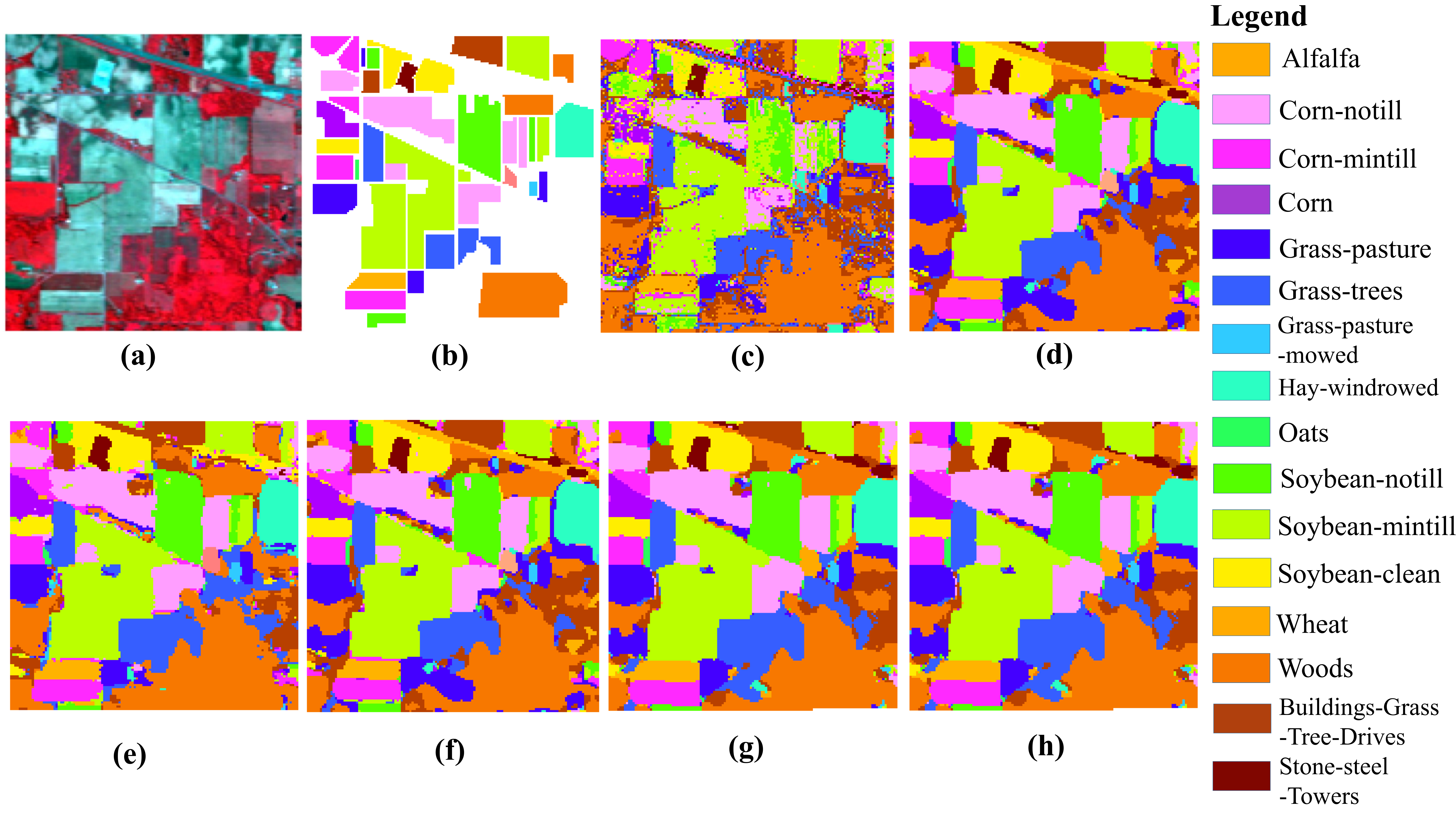}   %
    \caption{A comparison of the classification maps of IP data from six models: (a) the false colour composition map of the raw data, (b) Ground-truth data used in the training and evaluation of the models, (c-h) the classification results of DFCNN, VCNNs, SSCNN, SSRN, CapsNet, and the proposed model, respectively.}    %
    \label{fig:5}  %
\end{figure}

Fig. \ref{fig:5} demonstrates the classification maps of the proposed model and five competitors which are corresponding to optimal patch size configurations listed in Table \ref{table:3}. Because only the spectral signature of each pixel is used, there are noticeable \enquote{salt and pepper} noises found in the classification maps produced by the DFCNN and VCNNs (Fig. \ref{fig:5}c and \ref{fig:5}d). As a conventional neural network model, the SSCNN produces some misclassification in class boundaries (Fig. \ref{fig:5}e). The main reason is due to the typical defect of the convolutional operation, which makes the classification more sensitive to the spatial scale of the kernel size. The results from the SSRN, CapsNet and the proposed models do not show such noises and have the greater consistency with the ground truth data. Although similar classification maps are illustrated in Fig. \ref{fig:5}f-h for these three models, the classification map produced by our proposed model shows fewer misclassified pixels and clearer class edge and delineation. In addition, if we compare the labelled and unlabelled (not covered in Fig. \ref{fig:5}b) areas, there are less potential outliers in the resultant map of the proposed model. This indicates that the proposed model provides more consistent results on the task of ground surface classification than its competitors.\par

\begin{figure}[!t]   %
    \centering  %
    \includegraphics[width=3.5in]{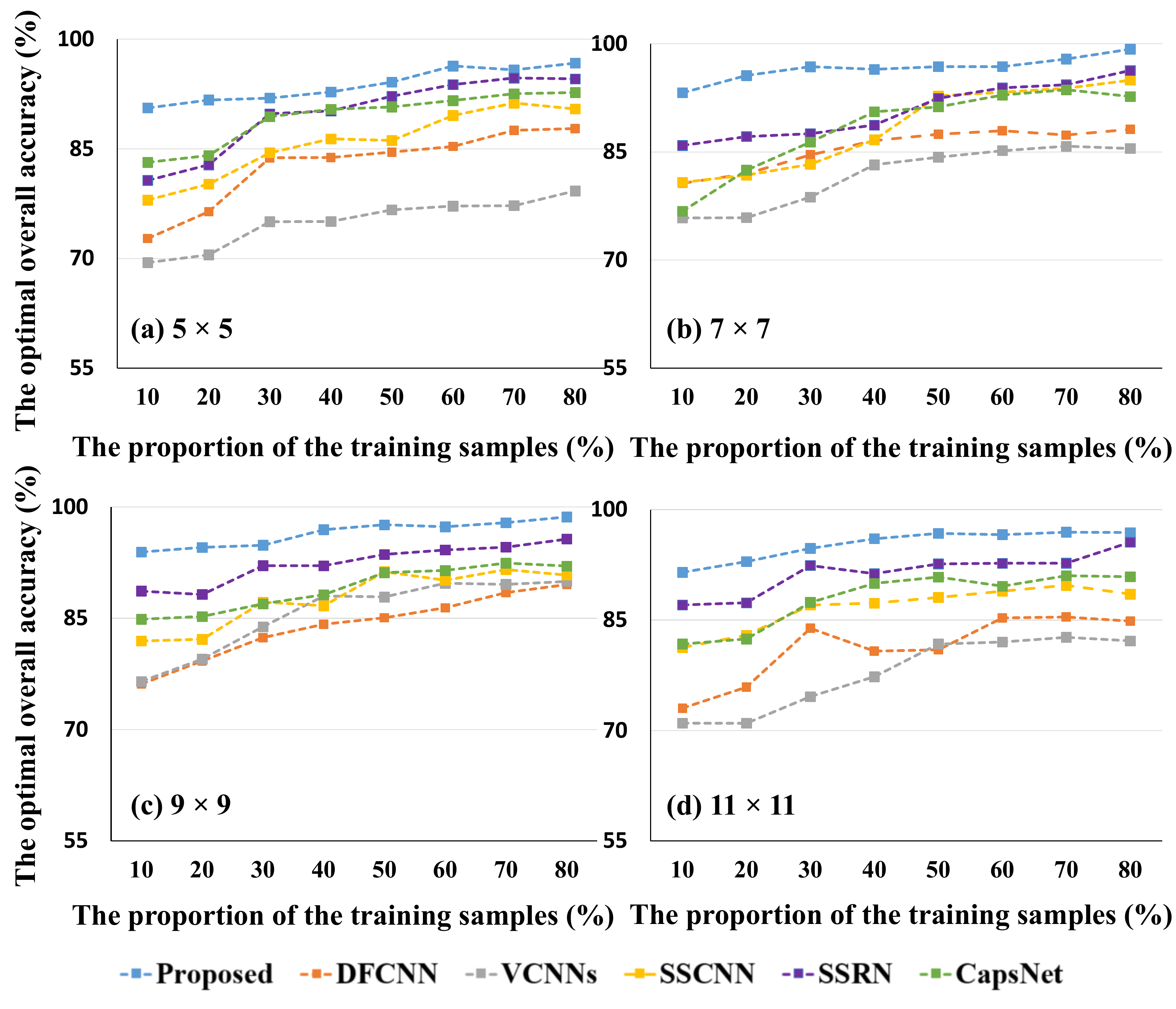}   %
    \caption{The relationships of sample size and overall accuracy for the proposed model and its five competitors on IP data with four different configurations of patch size (a) 5 $\times$ 5 (b) 7 $\times$ 7, (c) 9 $\times$ 9, (d) 11 $\times$ 11. The sample size varies from $10\%$ to $80\%$ of the total number of all labelled pixels}    %
    \label{fig:Trainaccessment-IP}  %
\end{figure}

Finally, we test the performance of the proposed model on small samples. Fig \ref{fig:Trainaccessment-IP} displays the relationship between the size of training sets and the overall accuracy for the proposed model and its competitors. The size of training sets varies from $10\%$ to $80\%$ of the number of all labelled pixels. In general, the classification accuracy improves with the increased sample size for all the models and for the four different sized input patches. The proposed model achieves the highest classification performance for all the four patch size configurations, and is less affected by the sample size. \par

\subsubsection{Experiment Two} Land cover classification using the PC dataset.\par
\label{subsec:exp2}
In this experiment, we evaluate the performance of the proposed method on the land cover classification using the PC dataset. \par
Firstly, we test the performance of the proposed model on large samples. Table \ref{table:PC-AA} lists the overall accuracy (OA) and standard deviation (STD) of the proposed approach and the competitors using different sized input patches. The results show that the proposed model achieves the best accuracy. Compared with the best model among five competitor for each patch size configuration, the proposed model achieves an improvement up to $4.24\%$ for $5 \times 5$ input patch, $2.95\%$ for $7 \times 7$ input patch, $1.69\%$ for $9 \times 9$ input patch, and $2.52\%$ for $11 \times 11$ input patch, respectively. 

Meanwhile, the standard deviations produced by the proposed approach are substantially lower than those produced by its competitors. The spatial size effect of input patches on classification performance varies between different models. The optimal size is $7 \times 7$ for the proposed model,SSCNN and CapsNet, and $9 \times 9$ for the DFCNN, VCNNs and SSRN. In general, all the models are not too sensitive to the spatial size of input patches. We also compared the storage and time costs between the proposed models and its five competitor. As shown in Table \ref{table:PC-trainsample}, the proposed model has the second best performance in running speed.    \par

Secondly, we further analysed the classification performance of the proposed model on each class in details. Table \ref{table:suplement-1} provides detailed classification accuracy of each land cover type for the proposed approach and its five competitors. Compared with its competitors, the proposed model achieves the best classification accuracy on the classes with biological properties (such as trees, meadows, and bare soil), and a comparative performance on non-biological land cover types (such as Bricks, Asphalts, Bitumen, Times, Shadows, Waters). The potential reason behind is that the biological information hidden in the HSI data is enhanced in Stage 1 of the proposed model. The OA and AA of the proposed model reaches $90.09\%$ and $90.3\%$, which outperforms all other five competitors. This is further confirmed by the results of McNemar's chi-squared ($\chi^2$) test, as shown in Table \ref{table:t2_stest}. The improvement of the proposed model in overall accuracy against all other competitators is statistically significant, with $\chi^2=32.49 (p \leq 0.01)$ for the proposed vs. DFCNN,  $\chi^2=22.41 (p \leq 0.05)$ for the proposed vs. VCNNs, with $\chi^2=20.58 (p \leq 0.05)$ for the proposed vs. SSCNN, with $\chi^2=12.01 (p \leq 0.1)$ for the proposed vs. SSRN, and with $\chi^2=10.96 (p \leq 0.1)$ for the proposed vs. CapsNet, respectively. \par

\begin{figure}[!t]   
\centering  
\includegraphics[width=3.5in]{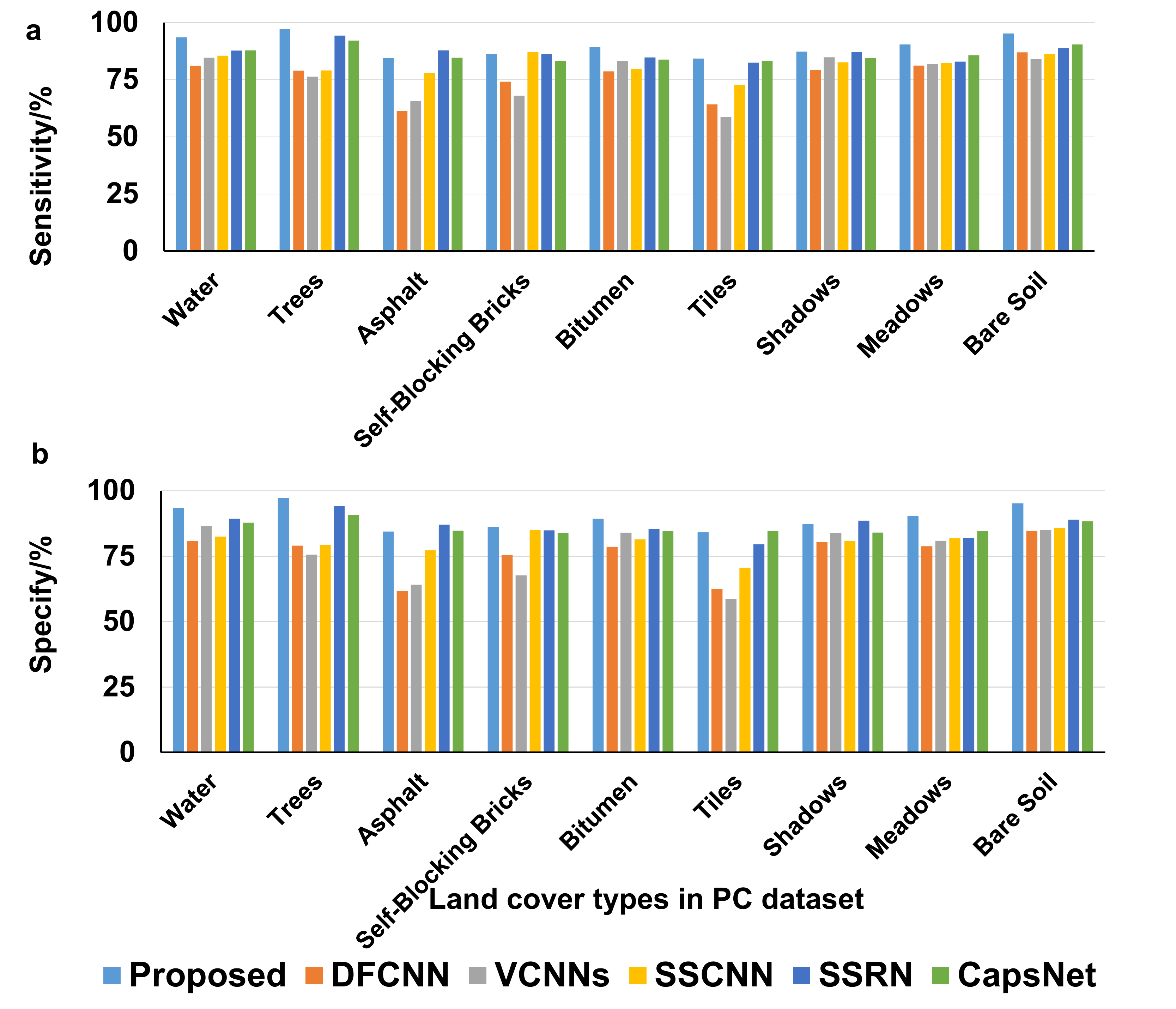}   
\caption{A comparison of the sensitivity and specificity of each of 9 classes in the multi-class plant species classification of the PC dataset from six models based on the optimal patch size configuration for each model (a) the sensitivity of each class, (b) the specificity of each class.}    
\label{fig:PCsensi}   
\end{figure}

Fig. \ref{fig:PCsensi} illustrates a detailed comparison of the sensitivity and specificity of each land cover type in the classification of the PC data from six models. Similar to the average accuracy, the proposed approach achieves the highest sensitivity and specificity on the classes with biological properties among 9 land cover types. This indicates that the proposed approach outperforms its competitors in reducing the leakage and misclassification of the biological information extraction and classification. \par

\begin{figure}[!t]   
\centering  
\includegraphics[width=3.5in]{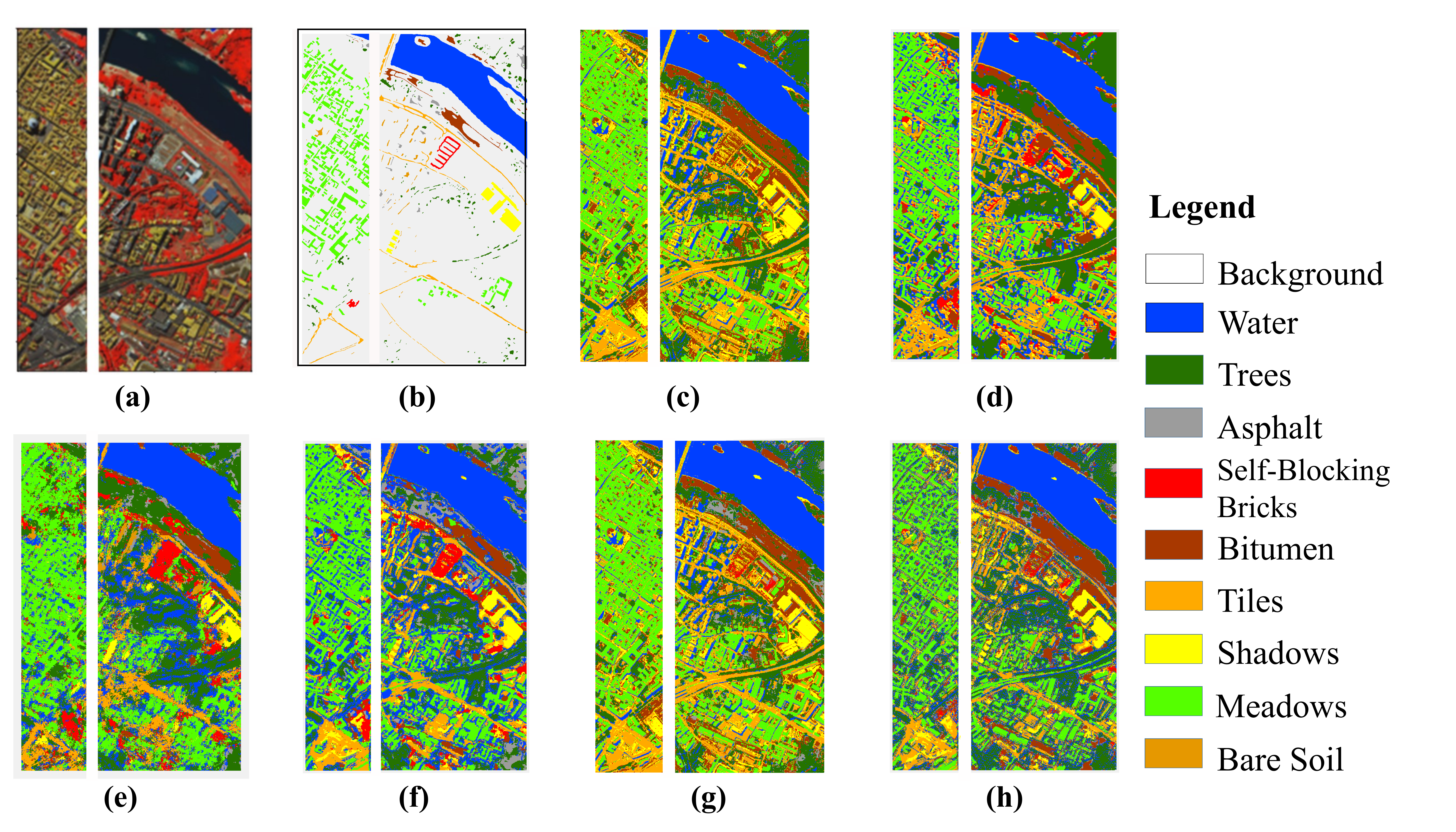}   
\caption{A comparison of the classification maps for PC data from six models (a) the false colour composition map of the raw data, (b) Ground-truth data used in the training and evaluation of the models. (c-h) the classification results of DFCNN, VCNNs, SSCNN, SSRN, CapsNet, and the proposed model, respectively.}    
\label{fig:PCmap} 
\end{figure}

Fig. \ref{fig:PCmap} displays the classification maps of all six models based on the optimal size configuration of input patches for each model (see Table \ref{table:suplement-1}). We can find that the DFCNN, VCNNs and SSCNN produce some noises and misclassification near class boundaries (see Fig. \ref{fig:PCmap}e). This may be due to the spatial convolutional operation, which makes the classification more sensitive to the kernel size. The classification maps from the SSRN, CapsNet and the proposed model (see Fig. \ref{fig:PCmap}f-h) show greater consistency with the ground truth data. If we only observe the three classes, trees, meadows and bear soil, we can find that the classification map from the proposed model has less outliers. \par

\begin{figure}[!t]   
\centering  
\includegraphics[width=3.5in]{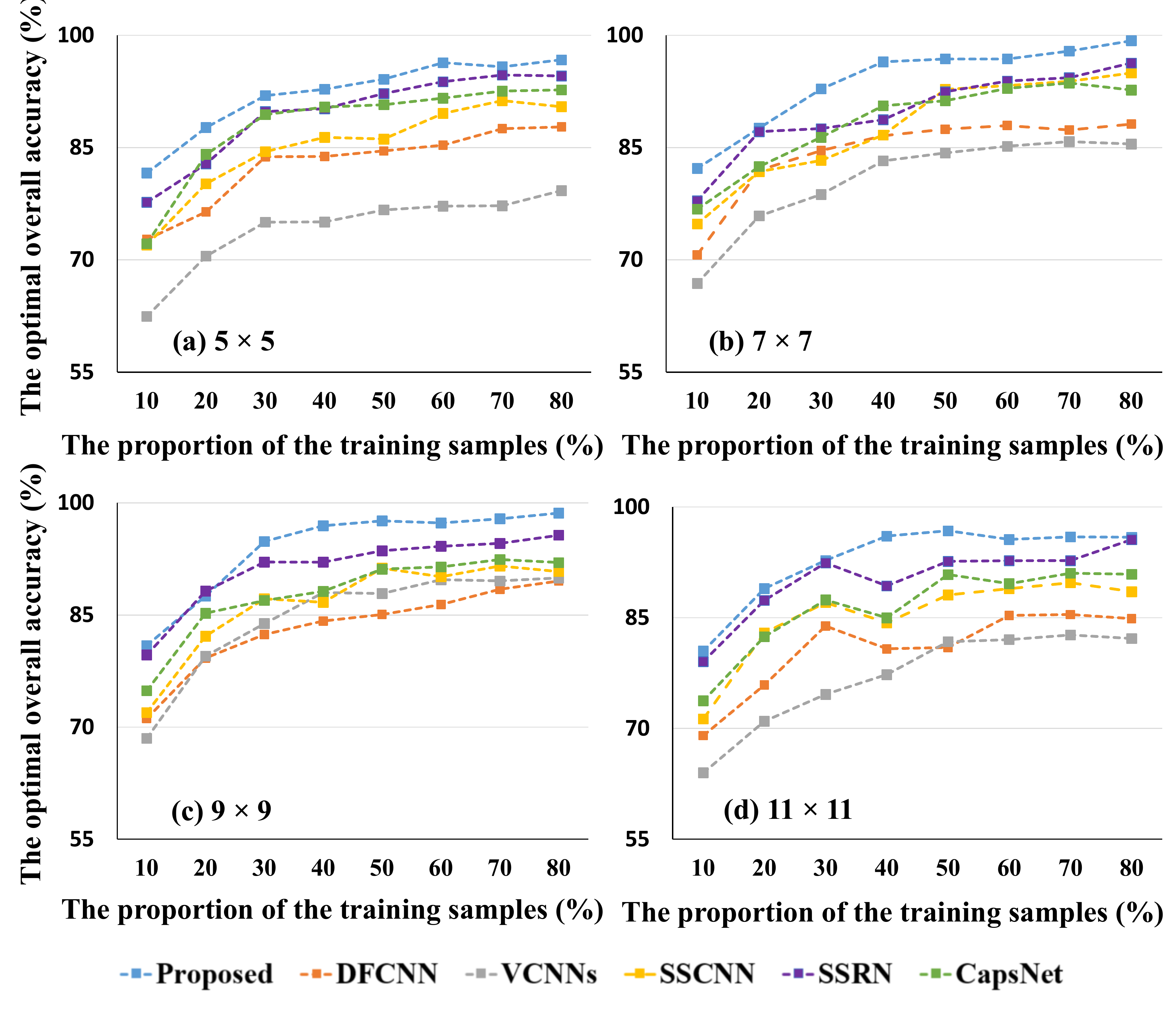}   
\caption{The relationships of sample size and overall accuracy for the proposed model and its five competitors on PC data with four sized patches: (a) 5 $\times$ 5, (b) 7 $\times$ 7, (c) 9 $\times$ 9, (d) 11 $\times$ 11. The sample size varies from $10\%$ to $80\%$ of the total number of all labelled pixels.}    
\label{fig:Trainaccessment-PC} 
\end{figure}

Finally, we test the performance of the proposed model on small samples. Fig \ref{fig:Trainaccessment-PC} presents the relationships of training set size and overall accuracy for the proposed model and its competitors with four different size configurations of input patches. In general, the classification accuracy increases with the increase of sample size for all the models. For all four patch size configurations, the proposed model again outperforms all the competitors in terms of overall accuracy on land cover classification. It can also be observed that the proposed model can achieve more than $90\%$ overall accuracy after the size of training set increases to more than $40\%$ of the number of all labelled samples except for the patch size configuration of $5 \times 5$. 
\begin{table}[!t]
\caption{Overall accuracy (OA) and standard deviation (STD) of land cover classification of the PC dataset using six models with four different spatial size configurations for input patches (* the best size configuration for each model is highlighted in bold.)} 

\label{table:PC-AA}
\centering
\resizebox{3.2in}{!}{
\begin{tabular}{ccccccccc}
\toprule
         & \multicolumn{2}{l}{5 $\times$ 5} & \multicolumn{2}{l}{7 $\times$ 7} & \multicolumn{2}{l}{9 $\times$ 9} & \multicolumn{2}{l}{11 $\times$ 11} \\ 
Class    & OA(\%)      & Std(\%)     & OA(\%)      & Std(\%)     & OA(\%)      & Std(\%)     & OA(\%)        & Std(\%)     \\ \hline
Proposed & 88.44       & 4.28        & \textbf{90.30}      & 3.74        & 89.31               & 5.58        & 89.10      & 5.81       \\
DFCNN    & 71.11       & 8.21        & 73.87               & 6.77        & \textbf{76.69}       & 6.23        & 76.82      & 6.92        \\
VCNNs    & 69.61       & 9.75        & 73.31               & 8.11        & \textbf{76.58}      & 6.79        & 75.08      & 6.37        \\
SSCNN    & 80.15       & 6.60        & \textbf{82.04}      & 5.28        & 81.44               & 5.92        & 80.28      & 5.67        \\
SSRN     & 82.93       & 6.18        & 84.96               & 5.42        & \textbf{87.62}      & 4.71        & 86.58      & 5.02        \\
CapsNet  & 84.20       & 4.97        & \textbf{87.35}       & 5.83        & 85.87               & 4.88        & 85.61      & 4.49  \\ \bottomrule
\end{tabular}
}
\end{table}

\begin{table}[!t]
\caption{Execution time and the number of model parameters of six models for PC data based on the optimal patch size configuration for each model} 
\label{table:PC-trainsample}
\centering
\resizebox{3.2in}{!}{
\begin{tabular}{ccccccc}
\toprule
           & Proposed & DFCNN  & VCNNs  & SSCNN  & SSRN   & CapsNet \\\hline
Parameters & 411057   & 583845 & \textbf{309224} & 470769 & 610057 & 451862  \\
Time(s)    & 517.6    & 601.8  & \textbf{457.4}  & 513.8  & 671.6  & 583.2 \\\bottomrule

\end{tabular}
}
\end{table}

\begin{table}[!t]
\caption{Land cover classification of the PC dataset from six models based on the optimal patch size configuration for each model} 
\label{table:suplement-1}
\centering
\resizebox{3.2in}{!}{
\begin{tabular}{cccccccc}
\toprule
Class                & Proposed & DFCNN & VCNNs & SSCNN & SSRN  & CapsNet \\ \hline
Water                & \textbf{95.54}     & 82.52  & 87.27  & 85.51  & 89.97  & 90.21    \\
Trees                & \textbf{99.22}     & 80.41  & 77.80   & 81.25  & 94.44  & 92.45    \\
Asphalt              & 84.40      & 63.19  & 67.01     & 79.45  & \textbf{89.52}  & 86.47    \\
Self-Blocking Bricks & 86.24     & 75.98  & 68.25  & \textbf{87.25}  & 86.24  & 84.40     \\
Bitumen              & \textbf{89.35}     & 80.24  & 86.14  & 82.14  & 86.28  & 85.32    \\
Tiles                & 84.28     & 65.47  & 61.28  & 73.50   & 82.55  & \textbf{84.82}    \\
Shadows              & 87.37     & 81.22  & 85.17  & 82.65  & \textbf{88.84}  & 84.61    \\
Meadows              & \textbf{90.44}     & 81.41  & 82.28  & 83.44  & 84.50   & 87.45    \\
Bare Soil            & \textbf{96.27}     & 87.27  & 86.15  & 88.21  & 90.14  & 90.88    \\ \hline
OA(\%)               & \textbf{90.09}	  & 76.69	& 76.58 & 82.04  & 87.62  & 87.35     \\
AA(\%)               & \textbf{90.30}	  & 77.57  & 77.88  & 82.58	 & 87.98  & 87.37     \\
Kappa                & 0.831    & 0.482 & 0.716 & 0.855 & 0.841 & 0.811   \\  \bottomrule

\end{tabular}
}
\end{table}

\begin{table}[]
\caption{McNemar's chi-squared ($\chi^2$) test with associated probability values for evaluating the statistic significance of accuracy differences in class-wise predictions between paired models on PC dataset. $***$ = $P \leq 0.01$, $**$ = $P \leq 0.05$, $*$ = $P \leq 0.1$, $NS$ = not significant.} 
\label{table:t2_stest}
\centering
\resizebox{3.2in}{!}{
\begin{tabular}{cccccc}
\toprule
Class           & \makecell[c]{Proposed \\ vs. DFCNN}     & \makecell[c]{Proposed \\ vs. VCNNs}     & \makecell[c]{Proposed \\vs. SSCNN}     & \makecell[c]{Proposed \\vs. SSRN}     & \makecell[c]{Proposed \\vs. CapsNet} \\ \midrule
Water                & 28.44***          & 22.49**           & 17.31**            & 13.01*            & 12.59*               \\
Trees                & 34.13***          & 27.44**           & 25.62**            & 22.52**           & 21.64**              \\
Asphalt              & 24.37***          & 21.18**           & 13.03*            & 17.95**           & 12.84*              \\
Self-Blocking Bricks & 21.43**           & 18.87**           & 12.16*            & 12.54*            & 4.31*                \\
Bitumen              & 17.77**           & 12.88*            & 11.53*            & 5.74*           & 5.41*                 \\
Tiles                & 15.29**            & 13.77*           & 6.78*             & 4.17*           & 4.05*                 \\
Shadows              & 19.13**            & 13.32*           & 12.24*            & 4.05*           & 5.54*                 \\
Meadows              & 24.82**           & 22.31**           & 17.00**              & 11.12*            & 10.34*                \\
Bare Soil            & 28.61**            & 20.83**           & 12.86*            & 5.61*        & 4.10*                  \\ \hline
Overall              & \textbf{32.49***}   & \textbf{22.41**}  & \textbf{20.58**}   & \textbf{12.01*}    & \textbf{10.96*}              \\ \bottomrule
\end{tabular}
}
\end{table}


\subsubsection{Experiment Three} Urban scene recognition using the UH dataset.\par
In this experiment, we evaluate the performance of the Urban scene recognition using the UH dataset. \par
Firstly, we test the performance of the proposed model on large samples. Table \ref{table:UH-AA} lists the OA and STD of the proposed approach and its five competitors with four different size configurations for input patches. It can be found that the proposed model outperforms all the competitors for all the patch size configurations. Compared with the best model among the five competitors, CapsNet, the proposed approach achieves an improvement up to $4.59\%$ for $5 \times 5$, $3.14\%$ for $7 \times 7$, $4.71\%$ for $9 \times 9$ and $5.98\%$ for $11 \times 11$ patch size configurations, respectively. 
Meanwhile, the standard deviations produced by the proposed approach are lower than those produced by the competitors. The optimal size of input patches is $7 \times 7$ for the proposed model, DFCNN, SSRN and CapsNet, and $9 \times 9$ for VCNNS and SSCNN. In terms of execution time and storage cost, the proposed model has the second best performance for the storage cost and the third best performance for the time cost, as shown in Table \ref{table:UH-trainsample}.    \par

Secondly, we further analysed the classification performance of the proposed model on each class. As shown in Table \ref{table:suplement-2}, the proposed model achieves the best classification on vegetation associated classes, such as healthy grass, stressed grass, synthetic grass and trees, but does not perform the best on manual scence, particularly worse on road. However, the OA and AA of the proposed model reaches $92.73\%$ and $91.43\%$ the best among all of the competitors. This is further confirmed by the results of McNemar's chi-squared($\chi^2$) test shown in Table \ref{table:t3_stest}, which is to evaluate the statistic significance of the accuracy differences between two-paired models.  The results show that the improvement of our proposed model in overall accuracy against all other competitors, is statistically significant with $\chi^2=30.99 (p \leq 0.01)$ for the proposed vs. DFCNN,  $\chi^2=29.98 (p \leq 0.01)$ for the proposed vs. VCNNs,  $\chi^2=23.45 (p \leq 0.05)$ the proposed vs. SSCNN,  $\chi^2=17.32 (p \leq 0.05)$ for the proposed vs. SSRN, and $\chi^2=11.02 (p \leq 0.1)$ for the proposed vs. CapsNet  respectively.

\begin{figure}[!t]   
    \centering  
    \includegraphics[width=3.5in]{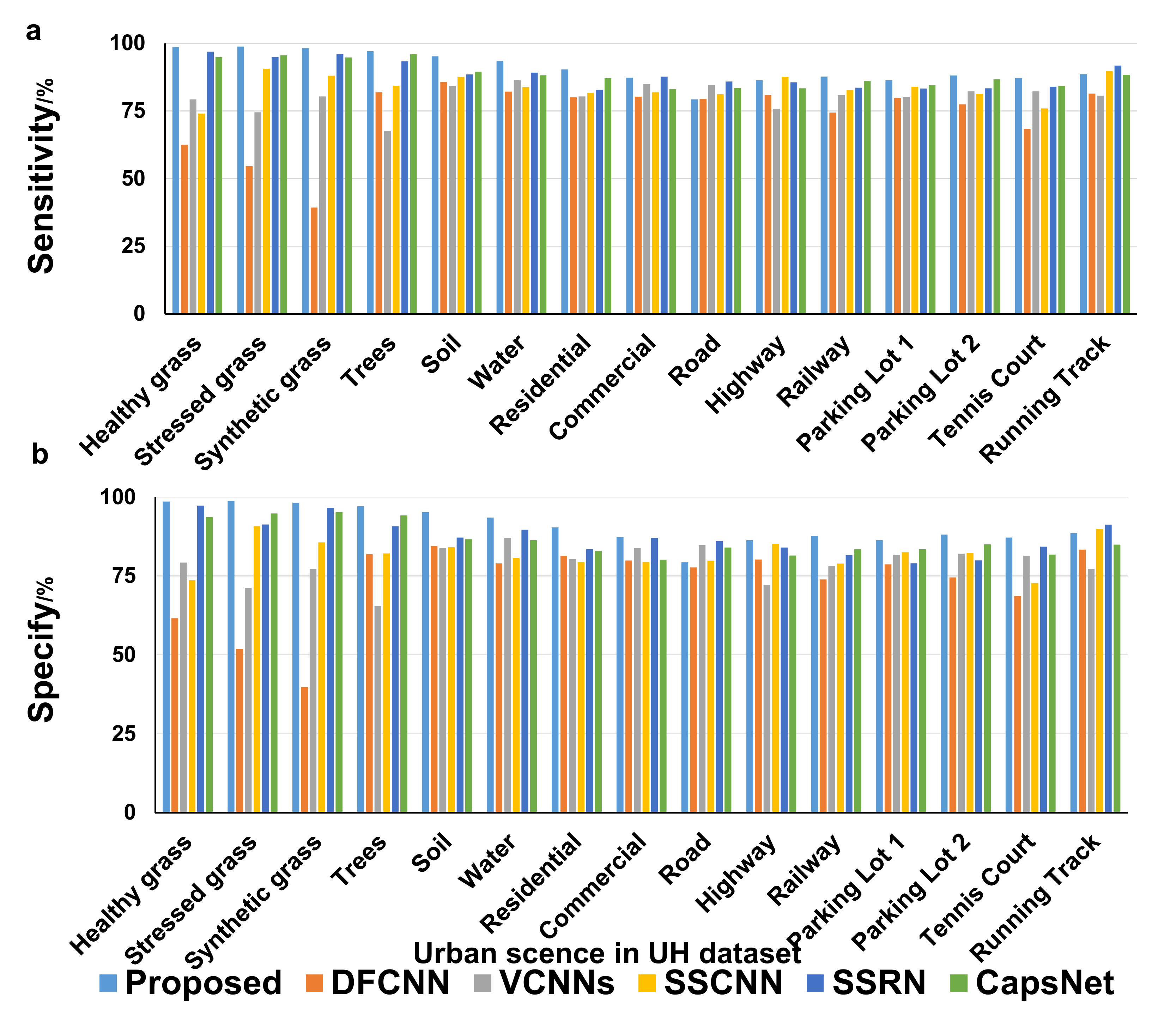}   
    \caption{A comparison of the sensitivity and specificity of each of 15 types in the urban scene recognition of the UH dataset from six models (a) the sensitivity of each type, (b) the specificity of each type.}    %
    \label{fig:UHsensi}  %
\end{figure}

Fig. \ref{fig:UHsensi} illustrates the detailed comparison of the sensitivity and specificity of each land cover type in the urban scene recognition of the UH dataset from the proposed model and the competitors. Similar to the classification accuracy, the proposed model achieves the highest sensitivity and specificity on the classes with biological properties, which further demonstrates that the proposed approach outperforms its competitors in reducing the leakage and misclassification of the biological information extraction and classification. \par

\begin{table}[!t]
\caption{Overall accuracy (OA) and standard deviation (STD) of the urban scene recognition of the UH data from six models with four different spatial size configurations as the input patches (Note: The best size configuration for each model is highlighted in bold.)} 

\label{table:UH-AA}
\centering
\resizebox{3.2in}{!}{
\begin{tabular}{ccccccccc}
\toprule
         & \multicolumn{2}{l}{5 $\times$ 5} & \multicolumn{2}{l}{7 $\times$ 7} & \multicolumn{2}{l}{9 $\times$ 9} & \multicolumn{2}{l}{11 $\times$ 11} \\
Class    & OA(\%)   & Std(\%)     & OA(\%)      & Std(\%)     & OA(\%)      & Std(\%)     & OA(\%)        & Std(\%)     \\ \hline
Proposed & 90.52    & 5.21         & \textbf{92.73}      & 4.48        & 90.89      & 5.95        & 90.91      & 5.75        \\
DFCNN    & 72.31    & 8.60         &\textbf{73.08}       & 8.62        & 72.14       & 7.27        & 71.48      & 6.09        \\
VCNNs    & 79.61    & 6.84        & 79.61       & 6.40     & \textbf{81.06}       & 3.21        & 78.74      & 7.64        \\
SSCNN    & 82.15    & 6.08        & 83.58       & 5.81    & \textbf{83.94}       & 4.27        & 82.04      & 4.83        \\
SSRN     & 85.93   & 7.88        & \textbf{88.75}      & 6.73        & 86.18      & 5.82        & 84.43      & 5.66        \\
CapsNet  & 85.20    & 7.73       & \textbf{89.59}      & 5.17        & 85.89      & 4.19        & 84.93      & 5.02      \\ \bottomrule
\end{tabular}
}
\end{table}

\begin{table}[!t]
\caption{Execution time and the number of model parameters of six models on UH data based on the optimal patch size configuration for each model } 
\label{table:UH-trainsample}
\centering
\resizebox{3.2in}{!}{
\begin{tabular}{ccccccc}
\toprule
            & Proposed & DFCNN   & VCNNs  & SSCNN  & SSRN    & CapsNet  \\ \hline
Parameters & 711057   & 1034845 & 608612 & 910769 & 1110057 & 859224   \\
Time(s)    & 865.6    & 1011.6  & 746.6  & 813.8  & 971.6   & 983.2 \\ \bottomrule
\end{tabular}
}
\end{table}

\begin{table}[!t]
\caption{Urban scene classification of the UH dataset from six models based on the optimal patch size configuration for each model.} 
\label{table:suplement-2}
\centering
\resizebox{3.2in}{!}{
\begin{tabular}{cccccccc}
\toprule
Class           & Proposed & DFCNN & VCNNs & SSCNN & SSRN & CapsNet \\ \hline
Healthy grass   & \textbf{99.62}    & 62.55 & 80.12 & 75.21 & 92.15 & 91.47   \\
Stressed grass  & \textbf{98.81}    & 56.32 & 75.24 & 92.24 & 90.24 & 95.81   \\
Synthetic grass & \textbf{98.32}    & 41.22 & 82.12 & 88.17 & 92.71 & 92.02   \\
Trees           & \textbf{99.14}    & 82.75 & 69.25 & 84.51 & 93.62 & 96.11   \\
Soil            & \textbf{95.21}    & 87.25 & 86.15 & 88.25 & 90.17 & 90.82   \\
Water           & \textbf{93.53}    & 82.55 & 87.24 & 85.54 & 89.94 & 90.21   \\
Residential     & \textbf{90.14}    & 81.64 & 82.22 & 83.41 & 84.55 & 87.43   \\
Commercial      & 87.63    & 81.82 & 85.14 & 82.65 & \textbf{88.82} & 84.62   \\
Road            & 84.38    & 80.24 & 86.12 & 82.11 & \textbf{86.24} & 85.34   \\
Highway         & 86.44    & 81.28 & 76.18 & \textbf{88.17} & 86.23 & 84.35   \\
Railway         & \textbf{87.73}    & 75.41 & 81.21 & 83.54 & 85.54 & 86.87   \\
Parking Lot 1   & 86.44    & 81.25 & 82.18 & 84.65 & 83.81 & \textbf{86.61}   \\
Parking Lot 2   & \textbf{88.12}    & 78.54 & 84.24 & 82.44 & 84.55 & 87.41   \\
Tennis Court    & \textbf{87.32}    & 69.42 & 82.35 & 76.24 & 85.21 & 85.52   \\
Running Track   & 88.68    & 83.41 & 82.21 & 90.36 & 93.23 & \textbf{89.51}   \\ \hline
OA(\%)          & \textbf{92.73}	& 73.08	& 81.06	& 83.94	& 88.75 & 89.59   \\
AA(\%)          & \textbf{91.43}    & 75.04 & 81.46 & 84.5  & 88.47 & 88.94   \\
Kappa           & 0.841    & 0.582 & 0.716 & 0.811 & 0.81  & 0.813   \\ \bottomrule
\end{tabular}
}
\end{table}

\begin{table}[!t]
\caption{McNemar's chi-squared ($\chi^2$) with associated probability values and p-test for the classification accuracy differences of the UH dataset. $***$ means $P \leq 0.01$, $**$ means $P \leq 0.05$, $*$ means $P \leq 0.1$.}
\label{table:t3_stest}
\centering
\resizebox{3.2in}{!}{
\begin{tabular}{cccccc}
\toprule
Class           & \makecell[c]{Proposed \\ vs. DFCNN}     & \makecell[c]{Proposed \\ vs. VCNNs}     & \makecell[c]{Proposed \\vs. SSCNN}     & \makecell[c]{Proposed \\vs. SSRN}     & \makecell[c]{Proposed \\v. CapsNet} \\ \midrule
Healthy grass   & 32.64***          & 24.13***          & 22.21**          & 18.89**          & 17.16**              \\
Stressed grass  & 30.36***          & 24.23***          & 21.52**          & 17.17**          & 19.59**              \\
Synthetic grass & 29.1***           & 20.89**           & 22.38**          & 12.57*           & 18.2**               \\
Trees           & 38.05***          & 35.39***          & 28.21***         & 25.35***         & 20.04**              \\
Soil            & 33.05***          & 24.87**           & 17.85**          & 11.91*           & 11.41*                  \\
Water           & 15.92**           & 15.81**           & 18.07**          & 12.21*           & 12.69*                \\
Residential     & 16.14**           & 17.05**             & 9.34*            & 7.67*            & 4.62*          \\
Commercial      & 18.49**           & 18.39**           & 13.69*           & 10.81*            & 6.56*         \\
Road            & 26.29***          & 24.01***          & 23.01**           & 16.59**          & 11.5*                \\
Highway         & 14.65**           & 11.03*           & 8.51*             & 6.81*             & 6.15*          \\
Railway         & 21.81**           & 24.41**           & 18.84**           & 13.02*           & 4.49*          \\
Parking Lot 1   & 16.94**            & 5.88*            & 7.44*            & 4.84*              & 15.87*                 \\
Parking Lot 2   & 22.93**           & 24.69**           & 17.32**           & 11.93*           & 3.94*            \\
Tennis Court    & 21.64**           & 21.17**           & 20.71**           & 12.7*             & 6.95*               \\
Running Track   & 28.38**           & 22.23**           & 20.51**           & 11.86*            & 11.84*               \\ \hline
Overall         & \textbf{30.99***}   & \textbf{29.98***}          & \textbf{23.45**}           & \textbf{17.32**}           & \textbf{11.02*}               \\ \bottomrule
\end{tabular}
}
\end{table}

\begin{figure}[!t]  
    \centering  
    \includegraphics[width=3.5in]{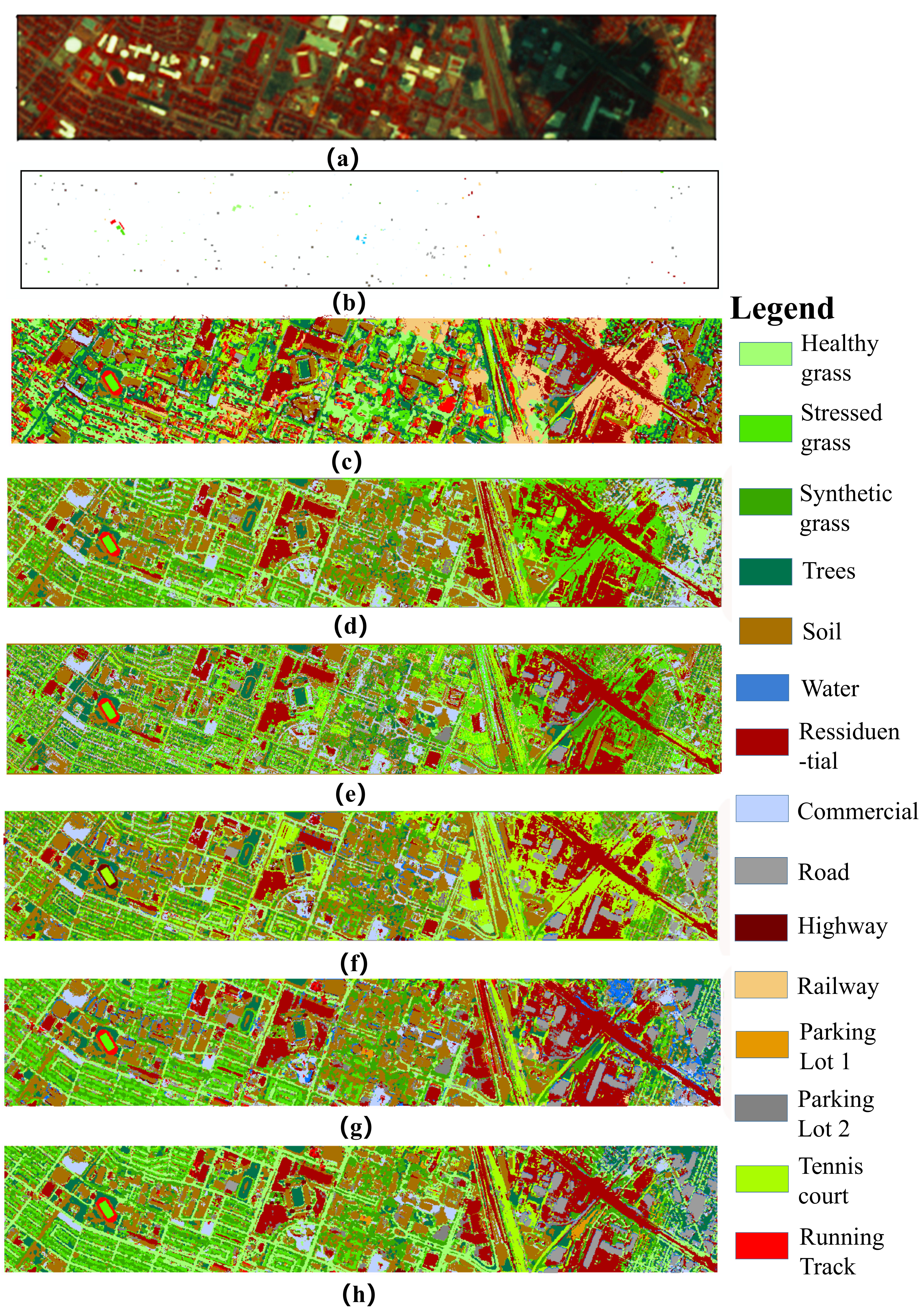}   
    \caption{The comparison of the classification maps of six models on the UH dataset: (a) the false colour composition map of the raw data, (b) Ground-truth data used in the training and evaluation of the models, (c-h) the classification result of DFCNN, VCNNs, SSCNN, SSRN, CapsNet, and the proposed model, respectively.}    
    \label{fig:UHmap}  
\end{figure}

Fig. \ref{fig:UHmap} demonstrates the classification maps of six models based on the optimal size configuration of input patches for each model (see Table \ref{table:suplement-2}). We can find some misclassification pixels located on the class boundaries produced by the DFCNN, VCNNs, and SSCNN (see Fig. \ref{fig:UHmap}c-e). The main reason behind may lie in the spatial convolutional operation which makes the classification more sensitive to the kernel size. The classification maps of the SSRN, CapsNet, and the proposed model (Fig. \ref{fig:UHmap}f-h) show  greater consistency with the ground truth data.

\begin{figure}[!t]   
    \centering  
    \includegraphics[width=3.5in]{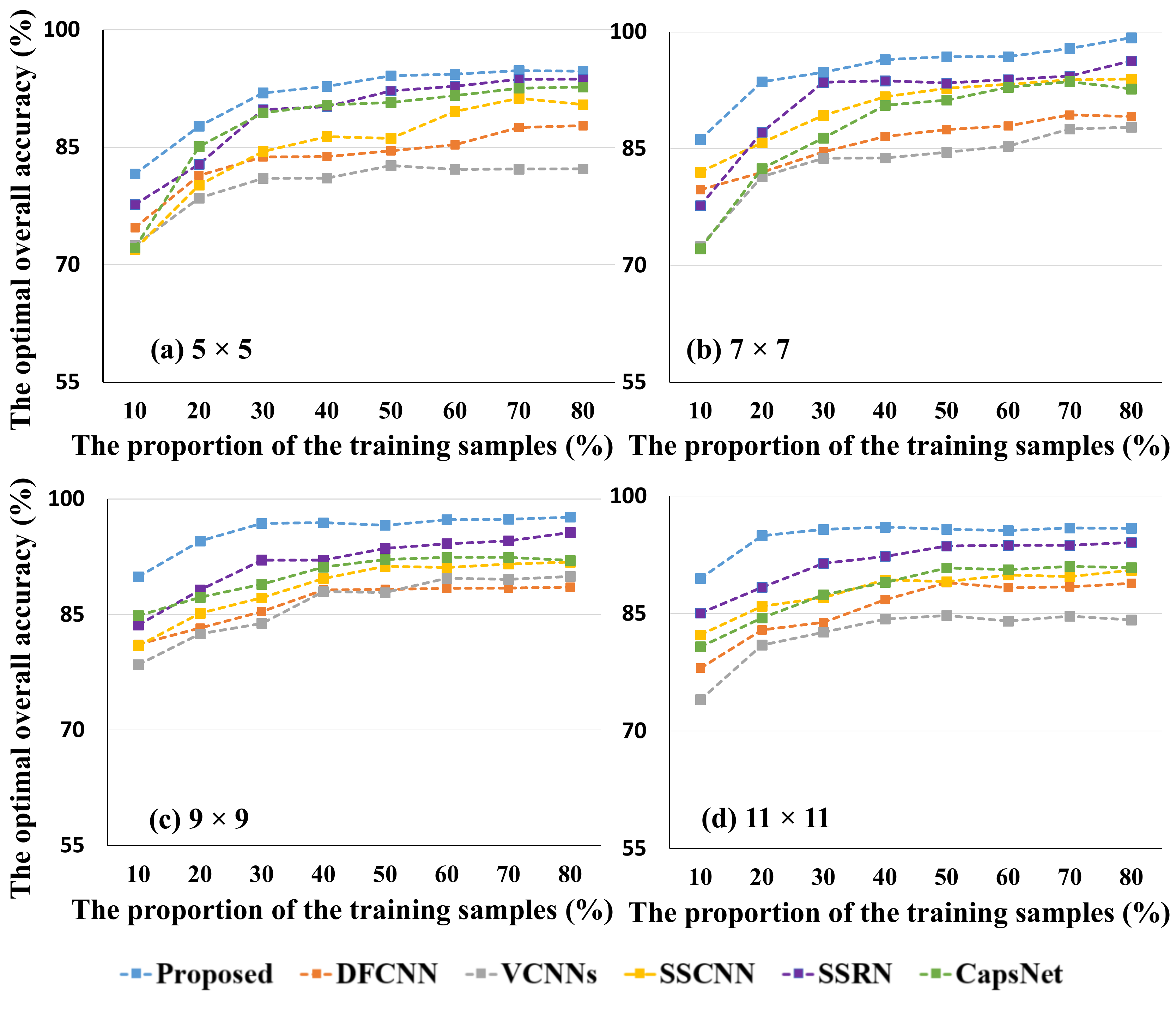}   
    \caption{The relationships of sample size and overall accuracy for the proposed model and its five competitors with four sized patches on the urban scene recognition of the UH dataset : (a) 5 $\times$ 5, (b) 7 $\times$ 7, (c) 9 $\times$ 9, (d) 11 $\times$ 11. The sample size varies from $10\%$ to $80\%$ of the total number of all labelled pixels.}   
    \label{fig:Trainaccessment-UH}  
\end{figure}

Finally, we test the performance of the propose model on small samples. Fig. \ref{fig:Trainaccessment-UH} displays the relationships of training set size with overall accuracy for all of the models with four different size configurations of input patches. Again, the proposed model outperforms all the competitors in terms of overall accuracy for all the four patch size configurations. It can also be observed that the proposed model can achieve more than $95\%$ overall accuracy after the size of training set increases above $30\%$ of the number of all labelled samples except for the size configuration of $5 \times 5$. In general, the overall accuracy increases with the increase of training set size for all six models.

\subsubsection{Experiment Four} Crop stress detection using the WYR dataset. \par
In this experiment, we evaluate the performance of the proposed method on crop disease diagnose using the WYR dataset.\par  
Firstly, we compare the OA of the proposed with its five competitors. As shown in Table \ref{table:WYR-AA}, the proposed model consistently outperforms its five competitors in terms of average accuracy for all the configurations of input patch size. 
The standard deviations produced by the proposed approach are lower than those produced by the competitors. In terms of execution time and storage cost, the proposed model have the second best performance for the storage cost and the third best performance for the time cost, as shown in Table \ref{table:WYR-trainsample}. The proposed model is not very sensitive to the path size, the size configurations $5 \times 5$, $7 \times 7$ and $9 \times 9$ provide almost the same accuracy. Compared with the best model among the five competitors, SSRN, the proposed approach achieves an improvement up to $4.1\%$ for $5 \times 5$, $3.34\%$ for $7 \times 7$, $6.34\%$ for $9 \times 9$ and $4.29\%$ for $11 \times 11$ patch size configurations, respectively. \par

Secondly, we further analysed the classification performance of the proposed model on each class in details. Table \ref{table:5} provides a detailed comparison of average accuracy of each of three types (i.e. healthy wheat, yellow rust and soil) in the crop disease detection of the WYR dataset from six models. The proposed model outperforms its five competitors in terms of average accuracy of each type and overall accuracy of all the three types. The OA and AA of the proposed model reaches $99.12\%$ and $98.89\%$ with a Kappa value of 0.84. 
This is further confirmed by the results of McNemar's chi-squared($\chi^2$) test, as shown in Table \ref{table:t4_stest}, which is to evaluate the statistic significance of the accuracy differences between two-paired models.  The results show that the improvement of our proposed model in overall accuracy against all other competitors, is statistically significant with $\chi^2=34.99 (p \leq 0.01)$ for the proposed vs. DFCNN, $\chi^2=32.78 (p \leq 0.01)$ for the proposed vs. VCNNs, $\chi^2=21.28 (p \leq 0.01)$ the proposed vs. SSCNN, $\chi^2=20.22 (p \leq 0.01)$ for the proposed vs. SSRN, and $\chi^2=18.72 (p \leq 0.01)$ for the proposed vs. CapsNet respectively. \par

\begin{table}[!t]
\caption{Overall accuracy (OA) and standard deviation (STD) of each class in the crop stress detection on the WYR dataset from six models based on the optimal patch size configuration for each model.  (Note: The best size configuration for each model is highlighted in bold.)}
\centering
\label{table:WYR-AA}
\resizebox{3.2in}{!}{
\begin{tabular}{ccccccccc}
\toprule
         & \multicolumn{2}{l}{5 $\times$ 5} & \multicolumn{2}{l}{7 $\times$ 7} & \multicolumn{2}{l}{9 $\times$ 9} & \multicolumn{2}{l}{11 $\times$ 11} \\ 
Class    & OA(\%)      & Std(\%)     & OA(\%)      & Std(\%)     & OA(\%)      & Std(\%)     & OA(\%)        & Std(\%)     \\ \midrule
Proposed & 96.52       & 3.54        & 96.88     & 2.42        & \textbf{99.12}     & 3.13        & 95.52     & 3.41        \\
DFCNN    & 81.11       & 5.03        & \textbf{81.73}       & 2.94        & 81.20        & 7.09        & 80.39      & 4.48        \\
VCNNs    & 77.45       & 5.88        & 78.80      & 5.32        & \textbf{78.91}       & 5.76        & 76.81      & 3.73        \\
SSCNN    & 85.25       & 7.17        & \textbf{86.85}       & 5.28        & 85.80        & 5.03        & 84.38      & 5.88        \\
SSRN     & 92.42       & 6.21        & \textbf{93.54}      & 5.51        & 92.02     & 4.33        & 91.23     & 4.49        \\
CapsNet  & 91.25       & 4.75        & 91.45      & 4.86        & \textbf{92.79}      & 3.77        & 90.99       & 3.63       \\ \bottomrule
\end{tabular}
}
\end{table}

\begin{table}[!t]
\caption{Execution time and the number of model parameters of six models on WYR data based on the optimal patch size configuration for each model } 
\label{table:WYR-trainsample}
\centering
\resizebox{3.2in}{!}{
\begin{tabular}{ccccccc}
\toprule
           & Proposed & DFCNN  & VCNNs  & SSCNN  & SSRN   & CapsNet  \\ \midrule
Parameters & 565075   & 985451 & \textbf{380214} & 767902 & 885407 & 792544   \\
Time(s)    & 865.6    & 1011.6 & \textbf{746.6}  & 813.8  & 971.6  & 983.2    \\\bottomrule

\end{tabular}
}
\end{table}

\begin{table}[!t]
\caption{Crop disease classification of the WYR dataset from six models}
\label{table:5} 
\centering
\resizebox{3.2in}{!}{
\begin{tabular}{ccccccc}
\toprule
Class         & Proposed & DFCNN & VCNNs & SSCNN & SSRN  & CapsNet  \\ \midrule
healthy wheat & \textbf{99.61}    & 89.55 & 84.17 & 92.24 & 96.15 & 95.48    \\
Yellow rust   & \textbf{98.85}    & 66.21 & 65.27 & 76.26 & 89.28 & 88.85    \\
Soil          & \textbf{98.21}    & 91.25 & 92.14 & 93.17 & 94.74 & 93.10    \\ \hline
OA(\%)        & \textbf{99.12}	  & 81.73 & 78.91 & 86.85 & 93.54 & 92.79        \\
AA(\%)        & \textbf{98.89}    & 82.34 & 80.53 &	87.22 &	93.39 &	92.48    \\
Kappa         & 0.841    & 0.582 & 0.716 & 0.811 & 0.810  & 0.813  \\ \bottomrule
\end{tabular}
}
\end{table}

\begin{table}[!t]
\caption{McNemar's chi-squared ($\chi^2$) with associated probability values and p-test for the classification accuracy differences of the WYR dataset. $***$ = $P \leq 0.01$, $**$ = $P \leq 0.05$, $*$ = $P \leq 0.1$.} 
\label{table:t4_stest}
\centering
\resizebox{3.2in}{!}{
\begin{tabular}{cccccc}
\toprule
Class           & \makecell[c]{Proposed \\ vs. DFCNN}     & \makecell[c]{Proposed \\ vs. VCNNs}     & \makecell[c]{Proposed \\vs. SSCNN}     & \makecell[c]{Proposed \\vs. SSRN}     & \makecell[c]{Proposed \\vs. CapsNet} \\ \midrule
Healthy wheat & 31.97***          & 26.74***          & 31.08***          & 26.24***          & 28.45***             \\
Yellow rust   & 35.92***          & 31.21***          & 26.98***          & 25.87***          & 24.61***             \\
Soil          & 33.6***           & 25.11***          & 18.09**           & 15.81**           & 15.36**              \\ \hline
Overall       & \textbf{34.99***}  & \textbf{32.78***}  & \textbf{21.28***}  & \textbf{20.22***}   & \textbf{18.72***}  \\ \bottomrule
\end{tabular}
}
\end{table}

\begin{figure}[!t]   %
    \centering  %
    \includegraphics[width=3.5in]{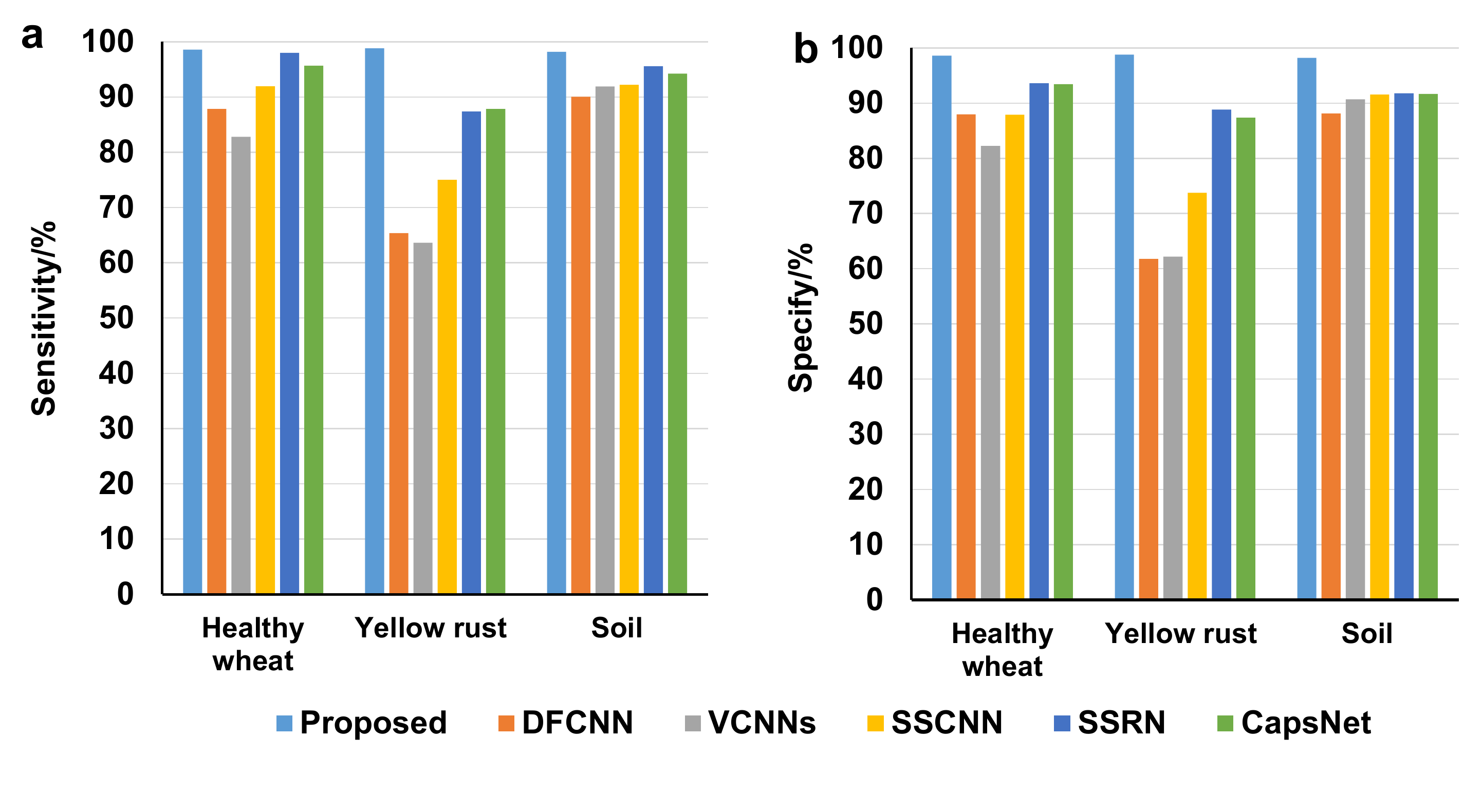}   
    \caption{ A comparison of the sensitivity and specificity of each detection type in the crop disease detection of the WYR dataset from six models: (a) The sensitivity of each class, (b) the specificity of each class.}    
    \label{fig:WYRsensi}  
\end{figure}

Fig. \ref{fig:WYRsensi} illustrates a detailed comparison of the sensitivity and specificity of each class from the optimal results of the proposed model and the competitors for the crop disease detection of the WYR dataset. Similar to the accuracy assessment results, the proposed model achieves the highest sensitivity and specificity on all three detection types, Healthy wheat, Yellow wheat and soil.

\begin{figure}[!t]   
    \centering  
    \includegraphics[width=3.5in]{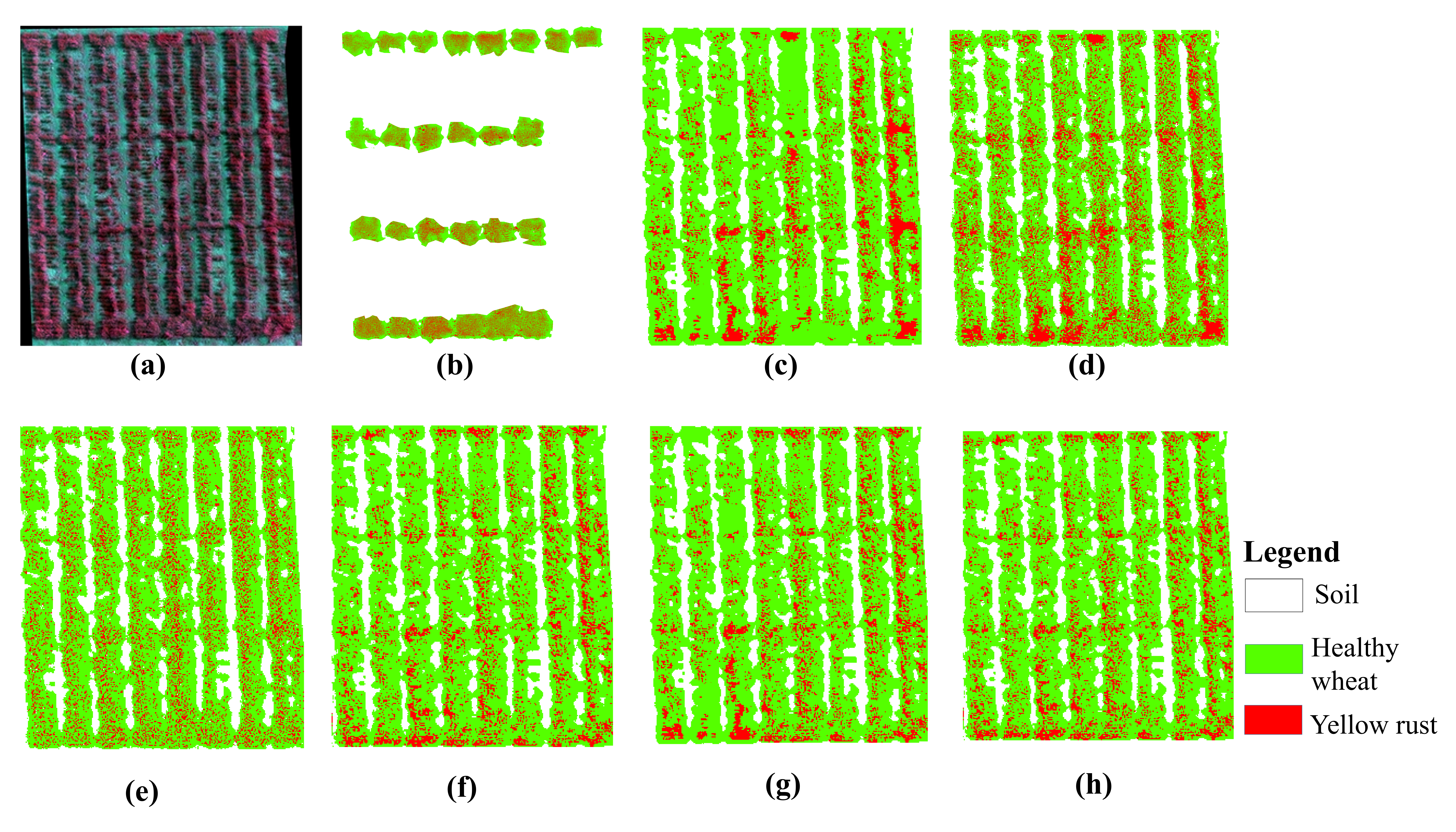}   
    \caption{A comparison of the classification maps of WYR dataset: (a) the false colour composition map of the raw data, (b) Ground-truth data used in the training and evaluation of the models. ($c-h$) the classification results of the FDCNN, VCNNs, SSCNN, SSRN, CapsNet, and the proposed model, respectively.}    
    \label{fig:7}  
\end{figure}

Fig. \ref{fig:7} shows the classification maps of yellow rust from a winter wheat field plot produced by the proposed approach and its five competitors with an optimal patch size for each model. The comparison of these four classification maps illustrates that the proposed approach outperforms the competitors in the class delineation and the distribution detection of yellow rust. More specifically, the class boundaries of pixels representing yellow rust obtained by the proposed approach are much clearer and more precise, such boundary characterizations are identical with the typical yellow rust pathogen features observed at the canopy scale. In addition, the yellow rust class contains stripe features, which have been better delineated in the maps obtained from the proposed approach. Moreover, if we look at the classification results over unlabelled areas, there is a noticeable consistency in classification with pathological distribution features, which suggests that the proposed approach provides a better generalization on the detection of wheat yellow rust than its competitors.

\begin{figure}[!t]  
    \centering  
    \includegraphics[width=3.5in]{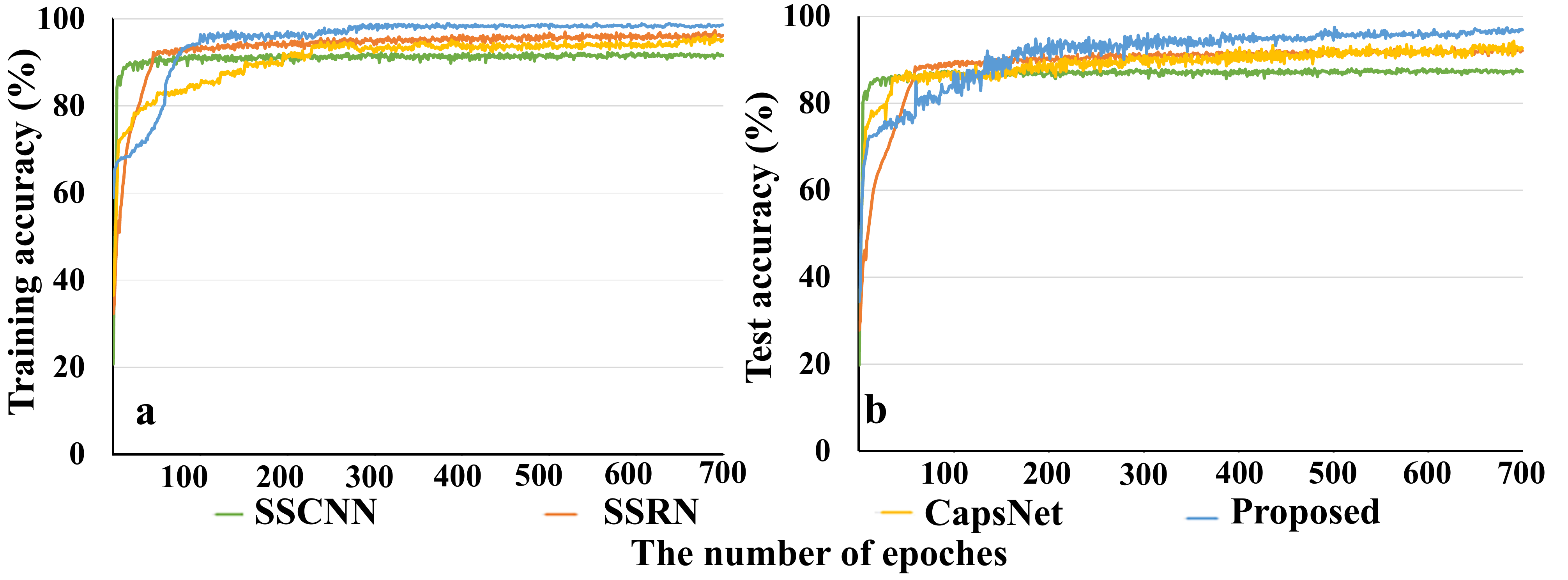}   
    \caption{Evolution of (a) training accuracy and (b) testing accuracy (in $\%$) of the SSCNN, SRNN, CapsNet and the proposed approach with a window size of $7 \times 7$  based on the WYR dataset.}    
    \label{fig:8}  
\end{figure}

For the purpose of demonstration, we present the convergence performance of the proposed model in this experiment. Fig. \ref{fig:8} shows the variations of the convergence of the proposed network architecture and its three competitors (SSCNN,SRNN and CapsNet) which have the same optimal patch size of $7 \times 7$. The epoch number is set to be 700. The results demonstrate that the proposed approach provides a more stable accuracy evolution in both training and testing processes. However, the accuracy decline can be observed in the training and testing processes of the competitors. For instance, the training OA of the SSCNN reaches $86.85\%$, but its testing accuracy only reaches $83.24\%$. The possible reason for such an accuracy decline phenomenon is due to the black-box learning process in the intermediate layers, which always leads the traditional network architecture to a local optimum. This occurs when the scale of sampling is not big enough to cover all of the possible states of the target classes. In this case, benefiting from the physical mechanism and interpretability as being discussed in previous sections, the learning process of the proposed method is able to represent the biophysical and biochemical variations and the spatial structure characteristics between the healthy wheat and the wheat infected with yellow rust. This explains why the proposed approached provides a greater performance in classification accuracy and robustness than the other models. \par
Regarding the convergence, it is noteworthy that the training OA of the proposed approach (the blue line in Fig. \ref{fig:8}a) reveals an S-shape curve, and we can separate this progress into four parts. At the beginning, the rate of convergence is slow during the first $50^{th}$ epochs. And then, this rate increases dramatically from $51^{th}$ to $120^{th}$ epochs. After that, the training accuracy reveals a fluctuation between $95.2\%$ and $97.5\%$ from the $121^{th}$ to $330^{th}$ epochs. Finally, the accuracy is stabilized at around $99.12\%$. This tendency may be associated to the two-stage network architecture of the proposed model, the training and extraction of sensitive spectral features in the first stage would produce a chain reaction for the learning process of the second stage and the final accuracy. Similarly, although the convergence rate of the proposed approach in the testing process is faster than that in the training process, it is still slower than the other three methods. \par

\begin{figure}[!t]   
    \centering  
    \includegraphics[width=3.5in]{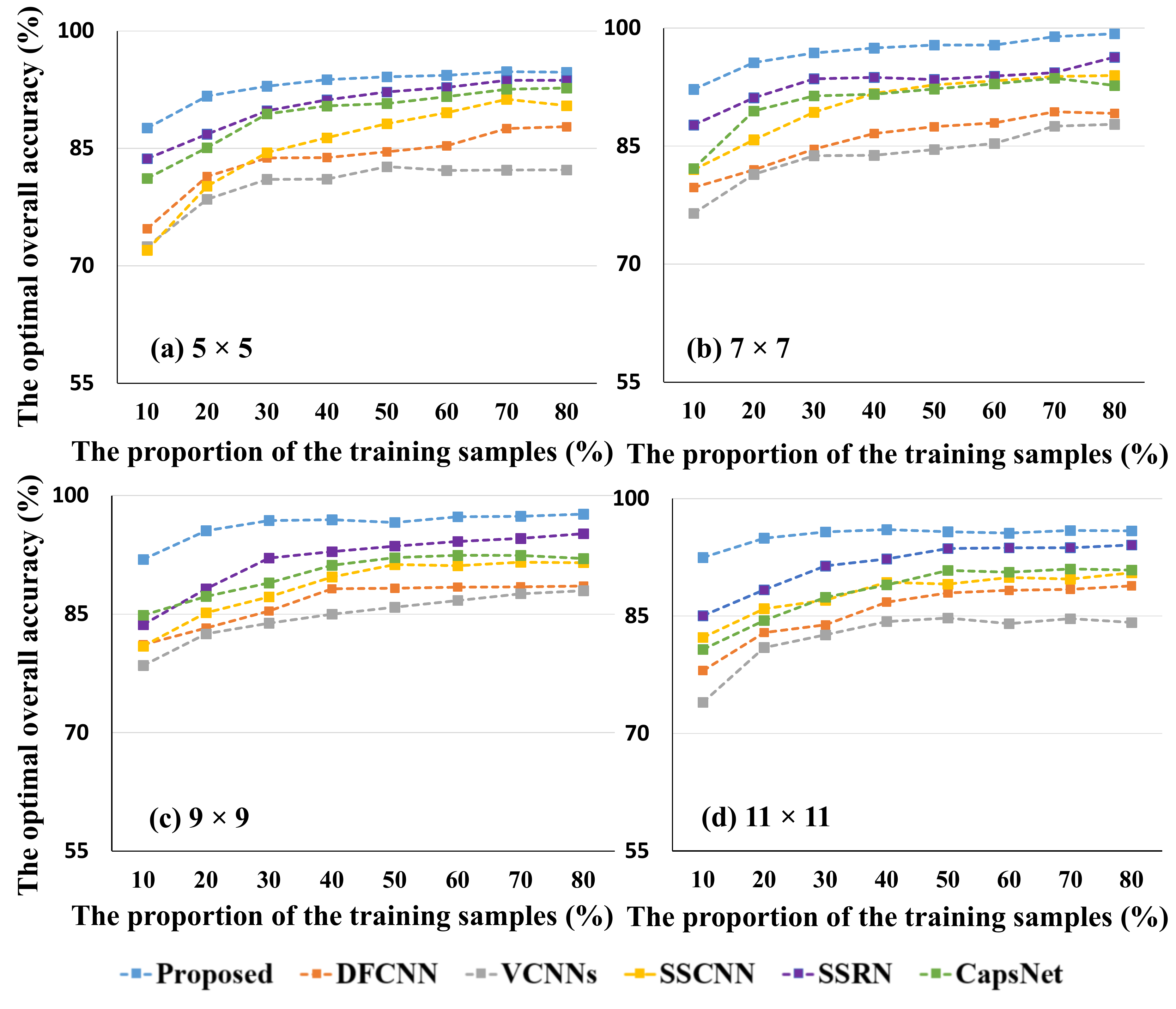}   
    \caption{The relationships of sample size and overall accuracy of the proposed model and its five competitors with four different sized patches: (a) 5 $\times$ 5 (b) 7 $\times$ 7, (c) 9 $\times$ 9 and (d) 11 $\times$ 11. The sample size varies from  $10\%$ to $80\%$ of the total number of all labelled pixels.}    
    \label{fig:Trainaccessment-WYR}  
\end{figure}

Finally, we test the performance of the proposed model on small samples.  Fig \ref{fig:Trainaccessment-WYR} presents the relationships of training set size and overall accuracy for the proposed model and its five competitors with four different size configurations for input patches on the crop stress detection of the WRY dataset. In general, the classification accuracy increases with the increase of sample size for all the models. The proposed model outperforms all the five competitors for all the four different sized input patches in terms of overall accuracy. \par

In summary, the proposed model outperforms all the five state-of-the-art deep learning models on five real-world datasets for vegetation information recognition in terms of average accuracy, overall accuracy, sensitivity and specificity, although it does not show big advantages in training speed for all the tasks. The proposed model also outperforms its competitors in reducing the leakage and misclassification during the biological information extraction and classification. Compared with its competitors, the proposed model is more robust to small samples. In general, the proposed model is not sensitive to the spatial size of input patches within the four chosen patch sizes, and has an optimal size of $7 \times 7$ for the four tasks in this study. \par

\section{Discussion}
\label{sec:discussion}

\subsection{The ablation analysis of the improved layers}

\begin{figure}[!t]   
\centering  
\includegraphics[width=3.4in]{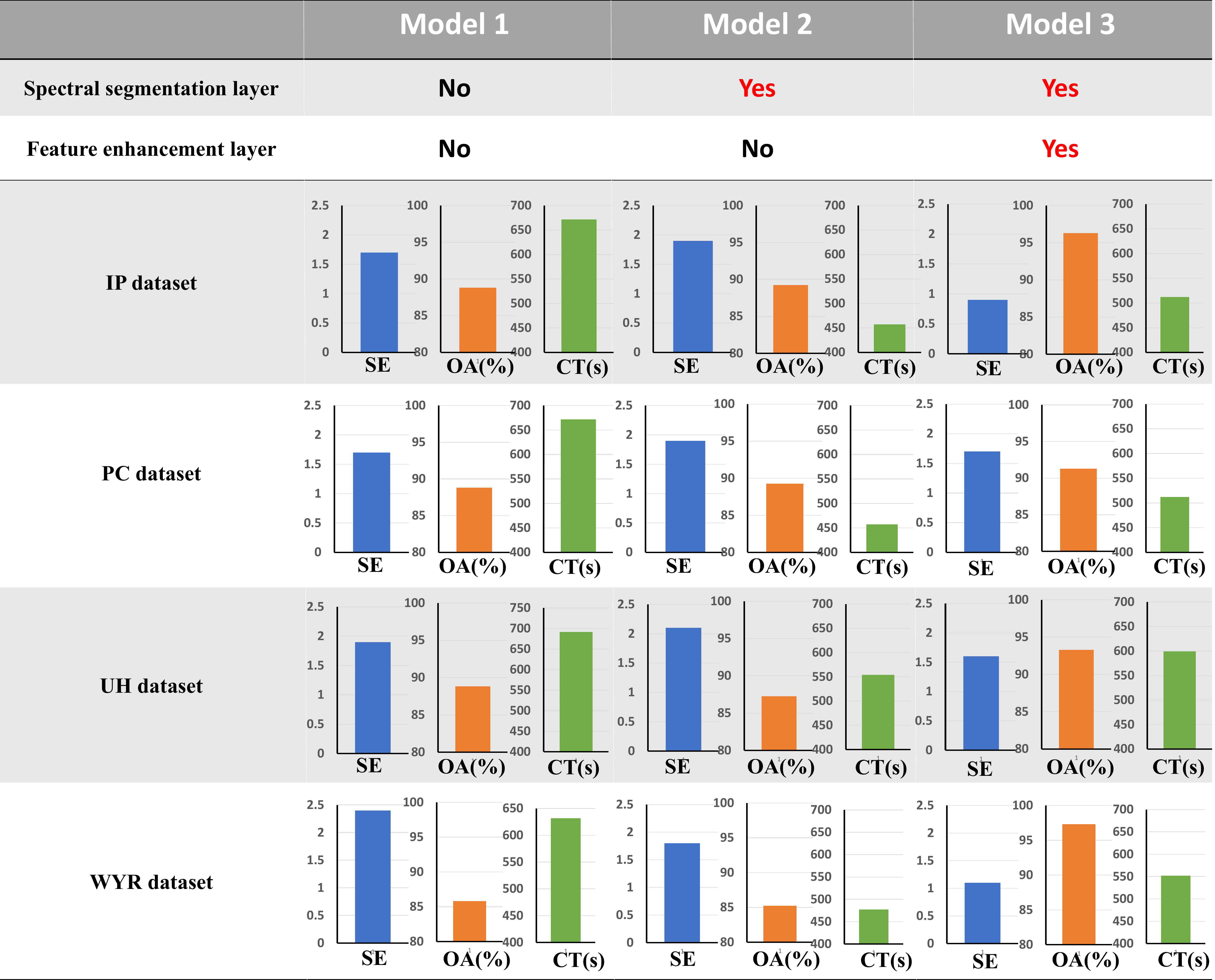}  
\caption{The results of an ablation analysis for evaluating the effects of the featured layers on vegetation information recognition from HSI data with four datasets. SE = Shannon entropy, OA = overall accuracy, and CT = computing time.}    
\label{fig:ablation}  
\end{figure}

In this study, two types of featured layers were used to improve the biological representations and classifications of deep learning models in the vegetation information recognition from HSI data: 1) spectral segmentation layer and 2) feature enhancement layer. To evaluate their effects on the model performance, we have conducted an ablation analysis that gradually introduced the two featured layers one by one into the base model, a DCNN-CapsNet joint network. Thus, there existed three different models (Fig. \ref{fig:ablation}), depending on how many different types of featured layers were included. We have evaluated their performance based on the Shannon entropy (a measure of uncertainty and disorder within the high level feature representations)\cite{sourati2018active}, overall accuracy, and computing time on four datasets.\par 

The base model without the featured layers is denoted as Model 1, in which HSI data with abundant spectral bands is directly input for feature extraction and classification. Differing from Model 1, Model 2 includs a spectral segmentation layers to split HSI input data into seven slices. This splitting operation makes the model pay more attention on the information clusters within the specific spectral slices rather than randomly perceived the class associated features across the redundant bands. Compared with Model 1, Model 2 achieves the similar Shannon entropy bound (1.7-2.4) and overall accuracy(86\%-89\%), but less computing time ($28.6\%$ reduction on average) on four datasets (Fig. \ref{fig:ablation}). This suggests that although the spectral segmentation process may not directly improve the feature representations for the classification, it improves the computing efficiency of the model. \par
 
As shown in the last column of Fig.\ref{fig:ablation}, Model 3 includes both segmentation and feature enhancement layers. The feature enhancement layer is introduced to enhance the homogeneous vegetation features between different spectral slices by allowing reliable feature inter-comparisons of photosynthetic activity and canopy characteristics. It can be found that the computing cost increases due to introducing the enhancement layer, but is still less than Model 1. In addition, compared to Model 1 and Model 2, Model 3 achieves lower Shannon entropy ($36.28\%$ lower on average) and higher overall accuracy ($12.18\%$ higher on average) on vegetation dominated datasets, IP and WYR. Although it is the same case on PC and UH datasets containing a number of non-vegetation classes, the improvement is less. The possible reason about this is due to the limited feature enhancement for the non-vegetation classes in these datasets from the enhancement layer. These results show that introducing the feature enhancement layer significantly suppresses the uncertainty and disorder of the feature space, and improves class representations for the classification decisions, especially for the vegetation classes.

\subsection{Interpretability analysis}
The interpretability of the model is one of the most important contributions in our work. We evaluate it from three perspectives: 1) pre-model interpretability, 2) post hoc analysis, 3) in-model interpretability with the proposed methods.
\subsubsection{Pre-model interpretability}
\begin{figure}[!t]   
\centering  
\includegraphics[width=3.5in]{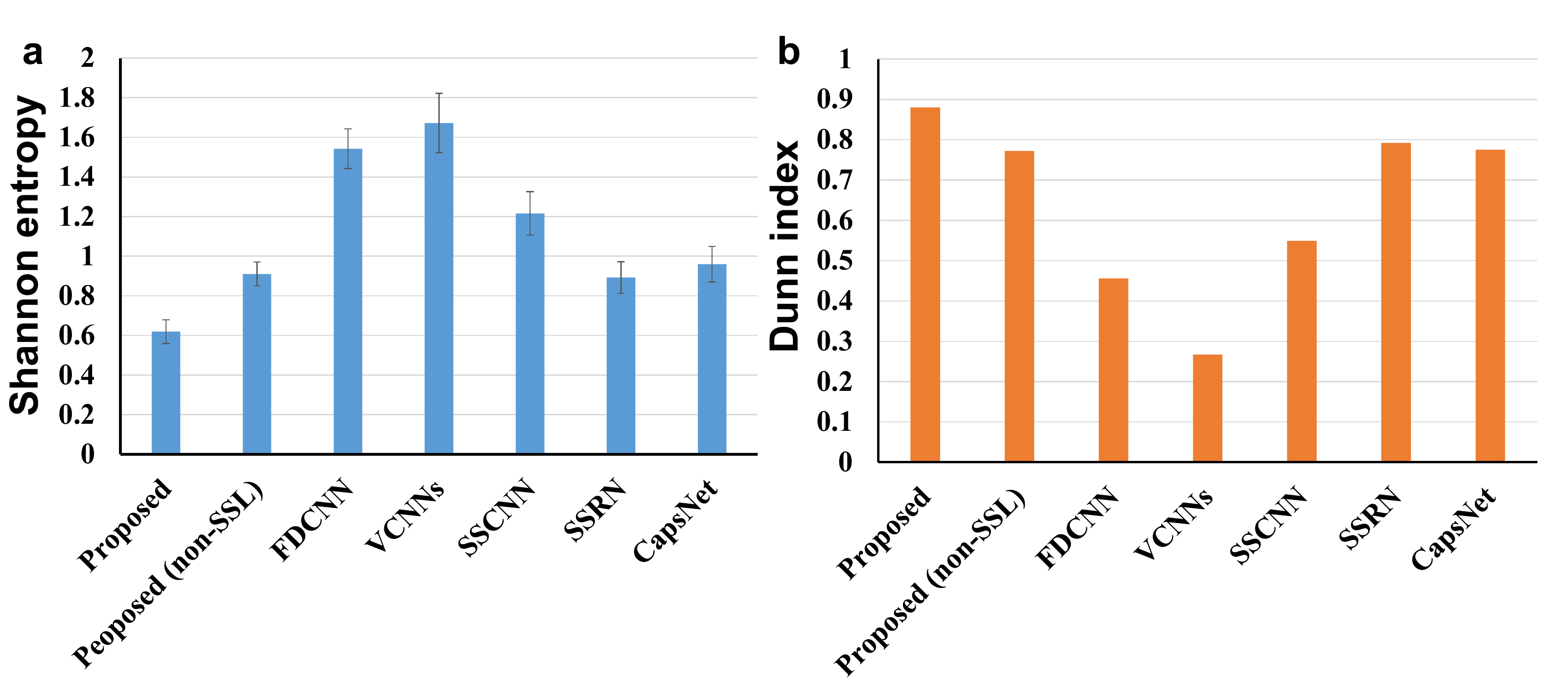}   
\caption{The Shannon entropy and Dunn index for all the classes in experiments $1 - 4$ . The values entropy is calculated for each class, then averaged all of them. The error bars represent the standard deviation across the classes}    
\label{fig:9}  
\end{figure}

The Shannon entropy and Dunn index are used to evaluate the effect of the spectral segmentation layer (SSL) on uncertainty and cluster of the features in the main model (see Fig. \ref{fig:9}). The Shannon entropy of each class is calculated by averaging all intermediate features within it, and the entropy values presented in Fig. \ref{fig:9} are the average of all classes in the Experiment one to four.  \par
The results indicate that, in comparison with the competitors, our proposed approach achieves a lower Shannon entropy (i.e. intra-class disorder) and a higher Dunn index (i.e. inter-class clustering). In addition, when we remove the spectral segmentation layer from the proposed model, the model performance has an observable decline. The rationale behind lies in that the spectral segmentation layer provides a physical constraint on the raw HSI data, which makes the learning process of the main model conduct in the band ranges with explicit spectral-biological attributes. Therefore, the intermediate features produced by the proposed approach may provide greater representations of the intrinsic inter-class differences and lower statistical-derived learning error than the other models.\par 

\subsubsection{ Post-model (post hoc) analysis}
\begin{figure}[!t]   
\centering 
\includegraphics[width=3.5in]{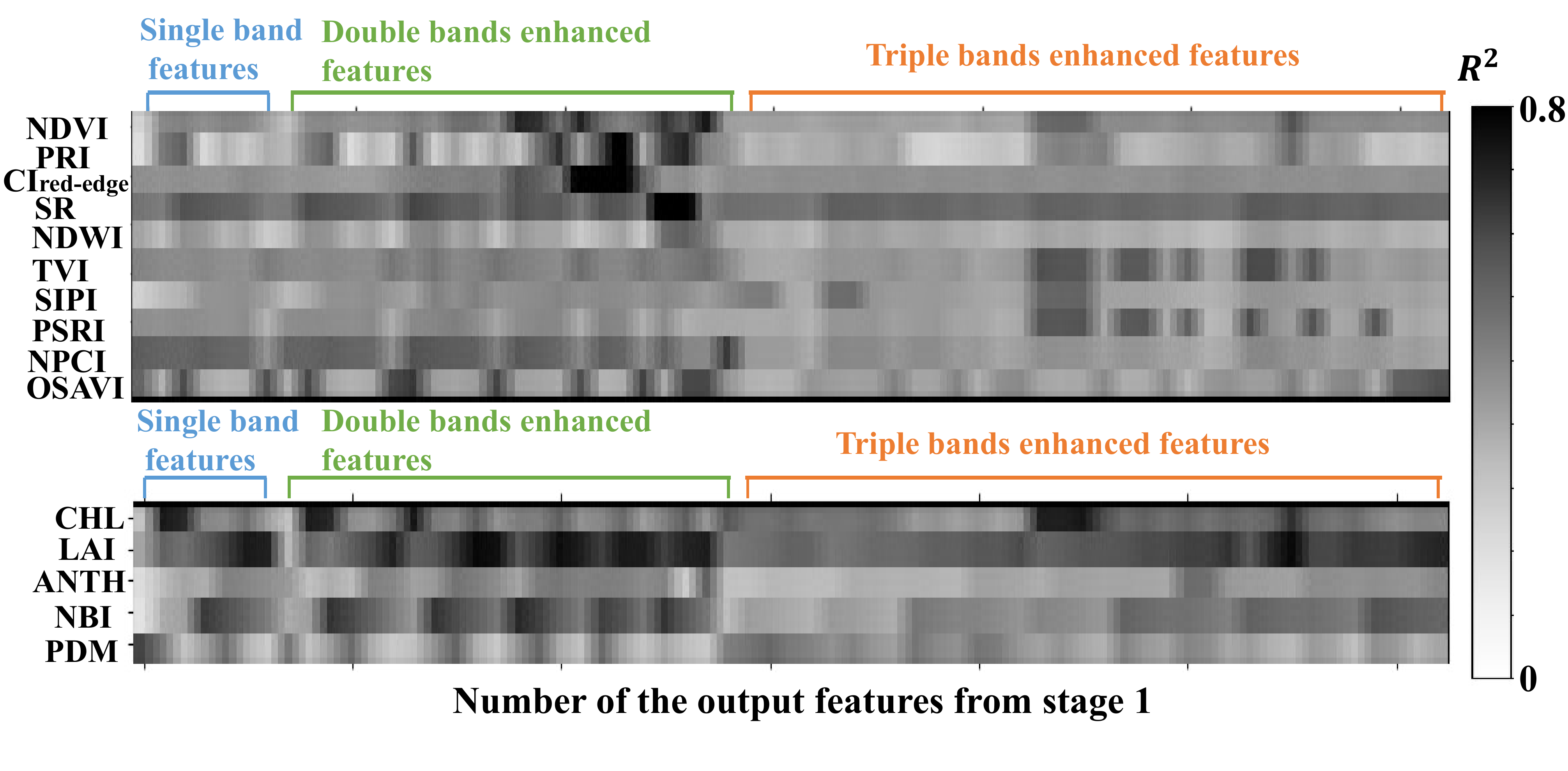}   
\caption{The visualisation of the correlation between the ground measured parameters and the extracted features from feature enhancement layer for (a) IP dataset and (b) WYR dataset}    
\label{fig:10}  
\end{figure}

In this work, the post hoc analysis mainly focuses on the enhanced interpretable feature block (i.e. stage 1) of the proposed network, the aim of the post hoc analysis is to explore the biological correlation between the intermediate spectral features with the auxiliary ground datasets. Considering the classes in PC dataset (Experiment Two) and UH dataset (Experiment Three) are non-biological entities, here, we just use the IP data (Experiment One) and WYR data (Experiment Four) as the study cases. For Experiment One, Fig. \ref{fig:10}a shows a correlation between the low-level features produced by the feature enhancement layers and the pre-selected vegetation indices, which reveals the potential spectral-derived biological properties of the enhanced spectral features. For instance, the red-edge and near-infrared slice derived features reveal the highest coefficient of determination ($R^2$) with the canopy structure associated SVIs, such as NDVI, PRI, and CIred-edge. Such correlations not only indicate the statistical representations of the generated spectral features, but also represent the subtle reflectance differences between the different plant categories in the Indian Pines dataset. Similarly, Fig. \ref{fig:10}b shows the correlation graph between the low-level features and the ground measured auxiliary parameters for the WYR dataset based disease detection task, which reveals the potential biophysical and biochemical properties hidden in the enhanced features. For instance, the red and red-edge slice derived features reveal the highest coefficient of determination ($R^2$) with the ground measured LAI, the green and red associated features exhibit a higher sensitivity with ground measured CHL. These findings from the two case studies based on the IP and WRY datasets have proven that the intermediate features can characterize not only the statistical properties for the corresponding classes, but also their biophysical and biochemical attributes, which provides the interpretability of the biological differences between the target classes. 

\subsubsection{In-model interpretability}
\begin{figure}[!t]  
\centering 
\includegraphics[width=3.5in]{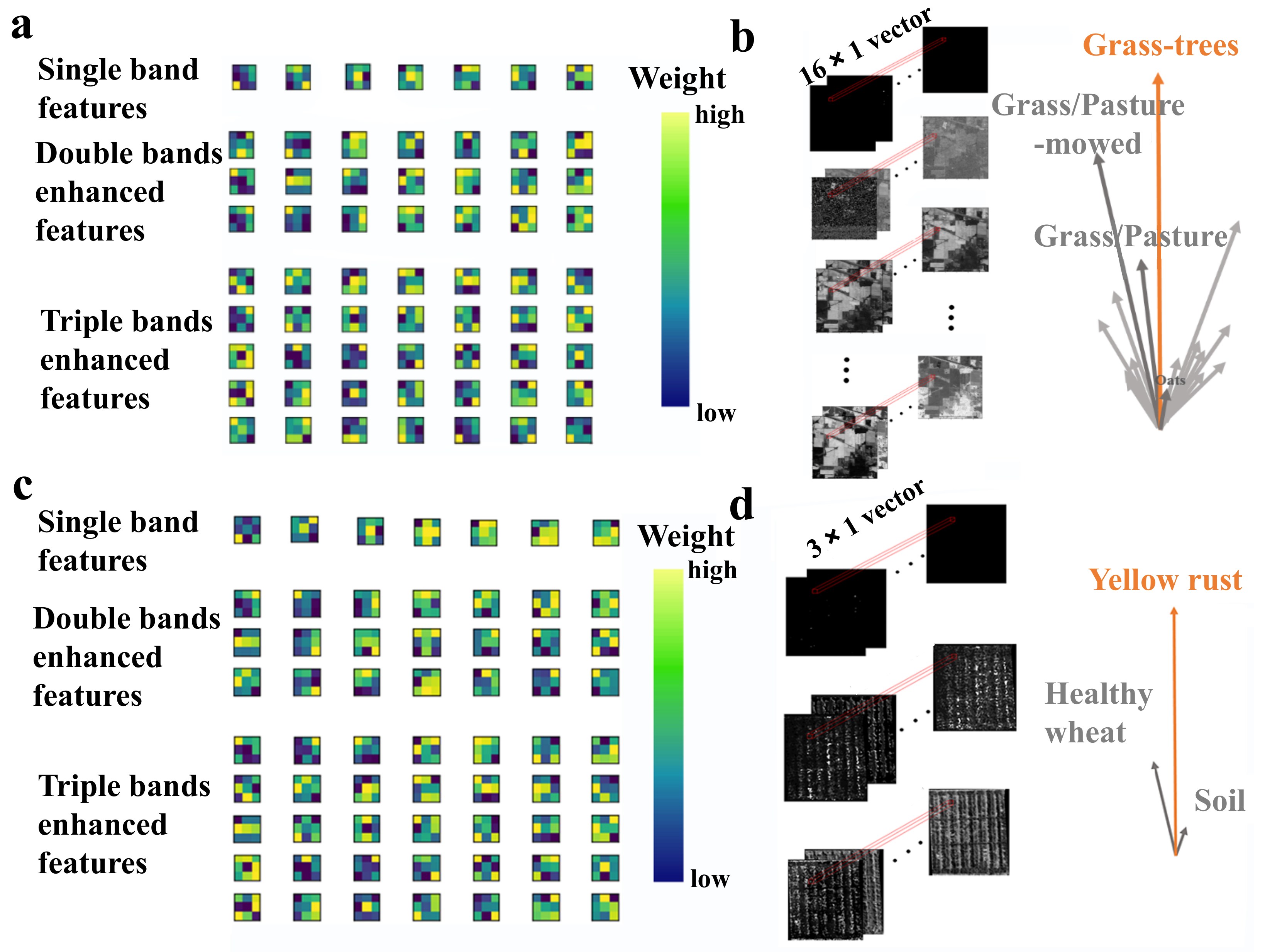}   %
\caption{The visualisation of the weights of $3 \times 3$ convolutional kernels in conv3 layer and the output feature maps and feature capsules of the capsule layers for Case Study 1: vegetation classification of IP datasets, (a-b) and Case Study 2: disease detection of WYR dataset, (c-d).}    %
\label{fig:11}  
\end{figure}

The main model splits the learning process into two stages: the spectral significance enhancement and spectral-spatial hierarchical construction representation. This logic explores the biophysical and biochemical hierarchical structure of the vegetation classes by encapsulating the extracted spectral-spatial information into capsule features. Besides, such an architecture improves the observation and explanation of the evolution of the features in different layers. Here, we analyse the biological interpretability of the high-level capsule features based on the IP dataset (Experiment One) and WYR dataset (Experiment Four). Fig. \ref{fig:11} illustrates the visualisation of the weights of the $3 \times 3$ convolutional kernels in $conv3$ layer, the output feature maps and feature capsules from the capsule layers for both IP and WYR datasets. This provides a direct way to understand the evolution progress of the intermediate features. Fig. \ref{fig:11}a visualises the $3 \times 3$ weight examples of the $covn3$ layer for IP data. It is noteworthy that the weights of the convolutional kernels for red, red-edge, and near-infrared associated features are higher than other features, which means the spectral features from red, red-edge, and near-infrared slices are more sensitive to the vegetation classes (e.g. the canopy structure characteristics in the IP dataset). These findings are also in agreement with previous studies \cite{RN55, RN61}. The final feature maps and the capsulized feature vectors from the capsule layers are shown in Fig. \ref{fig:11}b. where the well-designed capsule layer is able to manage the intermediate scalar features throughout the network, and also calculates the corresponding instantiation parameters to represent the hierarchical structure and potential transformations of the target classes. This will help better characterising the rotation invariance of spectral and spatial features of each class. The lengths of each feature vector are used to estimate the probability that a specific spectral-spatial feature occurred in each of the classes, and final classification would be determined by the maximum length.\par
For WYR data (i.e. Experiment Four), the visualisation of $3 \times 3$ weights of the covn3 layer (see Fig. \ref{fig:11}c) shows the weights of the neighbour pixels for the features sensitive to the biophysical parameters (e.g. LAI, PDM) are generally higher than the features sensitive to biochemical parameters (e.g. CHL, ANTH). This indicates that, comparing with the biochemical parameters, the texture information and spatial pattern provide more representations of the physical parameters in the detection and classification of yellow rust. In other words, the proposed approach provides better capability in characterizing the appearance symptom (e.g. leaf rolling, wither) when the wheat is infected by yellow rust. The feature maps from the class-capsules layer are shown in Fig. \ref{fig:11}d, the spectral-spatial features are integrated into three feature vectors, the length of each are used to estimate the probability that a specific biophysical and biochemical feature occurred in each of the classes, and final classification would be determined by the maximum length.

\section{Conclusion}
\label{sec:6}
In this study, a novel deep learning architecture based on two-stage spectral-spatial feature learning is presented for vegetation information recognition from HSI data with enhanced biological interpretability. The proposed model explores the potential biological and structural patterns of target entities from the inherent spectral-spatial information of the HSI data through feature mapping and transformation, and extracts the potential instantiation parameters (e.g. transformation and rotation) of these entities by means of a capsule network architecture, which enhances the performance and interpretability of the proposed model.  Specifically, the proposed network firstly splits the input HSI data into 7 spectral slices, and the most sensitive spectral features in each spectral slice are extracted. Subsequently, a set of enhanced features with the explicit biophysical and biochemical properties are generated based on the feature transformation rules (i.e. the binary index model and the triangular index model). Finally, a series of spectral-spatial capsule unites are employed to output the feature vectors that represent enhanced feature sets serving as a collection of canonical spectral-spatial patterns and the specific instantiation parameters at a higher level. Through this network, the intermediate features is capable of representing more biological and structural patterns of ground vegetation entities, which subsequently leads to increased model interpretability and a reduction of the computing complexity, and therefore, a more accurate model. An ablation analysis confirms these advantages. The comparison with five state-of-the-art models for HSI data classification reveals that the proposed BIT-DNN exhibits a very competitive performance in classification accuracy.\par

\section*{Acknowledgement}

This research is supported by BBSRC (BB/R019983/1), BBSRC (BB/S020969/1). The work is also supported by Newton Fund Institutional Links grant, ID 332438911, under the Newton-Ungku Omar Fund partnership (the grant is funded by the UK Department of Business, Energy, and Industrial Strategy (BEIS) and the Malaysian Industry-Government Group for High Technology and delivered by the British Council. For further information, please visit www.newtonfund.ac.uk.)

\ifCLASSOPTIONcaptionsoff
  \newpage
\fi



\bibliographystyle{IEEEtran} 
\bibliography{fefnew}
\end{document}